\documentclass[letterpaper, 10 pt, conference]{ieeeconf}  

\IEEEoverridecommandlockouts                              

\usepackage{graphics} 
\usepackage{graphicx}
\usepackage{epsfig} 
\usepackage{mathptmx} 
\usepackage{times} 
\usepackage{amsmath} 
\usepackage{amssymb}  
\usepackage{tabularx}
\usepackage{subcaption}
\usepackage{balance}
\usepackage{url}
\usepackage{float}

\title{\LARGE \bf
Demand-Driven Vertiport Siting and Discrete-Event Fleet Simulation for On-Demand Urban Air Mobility Network Design
}

\author{Hossein Z. Saghazadeh$^1$, Yonas Ayalew$^1$, \IEEEmembership{Member, IEEE}, Reza Ahmari$^2$,\\Parham Kebria$^1$, \IEEEmembership{Senior Member, IEEE}, and Abdollah Homaifar$^1$, \IEEEmembership{Senior Member, IEEE}
\thanks{*This work was supported by NCDOT under grant No. RP2025-43, the National Aeronautics and Space Administration (NASA) University Leadership Initiative (ULI) under grant numbers 80NSSC20M0161, 80NSSC25M7098 and in part by US-DOT under grant No. 69A3552348327, and by National Science Foundation (NSF) under grant No. 2301553.}
\thanks{$^1$The authors are with the department of Electrical and Computer Engineering, North Carolina A\&T State University, Greensboro, 27411 NC, USA. Corresponding author: {\tt\small homaifar@ncat.edu}}%
\thanks{$^2$The author is with the department of Computer Science, North Carolina A\&T State University, Greensboro, 27411 NC, USA.}
}

\begin{document}

\maketitle

\begingroup
\renewcommand{\thefootnote}{}
\footnotetext{\scriptsize
© 2026 IEEE.  Personal use of this material is permitted.  Permission from IEEE must be obtained for all other uses, in any current or future media, including reprinting/republishing this material for advertising or promotional purposes, creating new collective works, for resale or redistribution to servers or lists, or reuse of any copyrighted component of this work in other works.
}
\endgroup

\begin{abstract}
This paper presents a demand-driven framework for on-demand Urban Air Mobility (UAM) network design that links vertiport siting, fleet simulation, and door-to-door travel-time feasibility. Demand is estimated from commuter and passenger activity data, converted into spatial trip-end points, and clustered using $K$-means to generate candidate vertiport locations. Candidate networks are screened using range and minimum station-spacing constraints, then evaluated with a discrete-event simulation that models multi-vehicle dispatch, deadhead relocation, battery swaps, and service regularity. Flight time and energy consumption are computed using a point-mass eVTOL performance model. In a Greater Los Angeles case study, the preferred design expands from four stations and four eVTOLs at low demand to sixteen stations and twelve eVTOLs at the highest tested demand level. Results show that larger fleets improve completion time and vehicle-arrival regularity but do not eliminate deadhead flights, indicating that spatial demand imbalance remains an operational burden. The travel-time savings analysis further suggests that UAM is most defensible for longer or congestion-heavy trips where sufficient non-flight time remains after accounting for flight time.
\end{abstract}


\section{INTRODUCTION}
Urban Air Mobility (UAM) uses electric vertical take-off and landing (eVTOL) aircraft to provide on-demand trips between dedicated facilities, such as vertiports and vertistops, with ground access and egress. Recent concepts of operations describe a progression from low-tempo deployments to denser networks, making vertiport siting and fleet planning practical design problems \cite{FAA_ConOps_2_2023}. Prior studies show that scalable UAM depends not only on aircraft performance, but also on vertiport availability, facility processing, fleet management, and public acceptance \cite{Garrow_UAM_Review_2021}. Therefore, planning methods are needed that connect spatial demand, facility placement, aircraft performance, and network-level operations.

Existing UAM studies address demand estimation, vertiport siting, network design, and fleet operations. Demand-side studies often define addressable trips using door-to-door travel-time savings after access, egress, and processing times are included \cite{Bulusu_DemandAnalysis_2021,WuZhang_IntegratedDesignDemand_2021}. For siting, clustering methods provide scalable and interpretable candidate locations, while recent comparisons show that the clustering choice can affect predicted benefits and equity outcomes \cite{Jeong_Kmeans_Vertiports_2021,Guo_ClusteringComparison_2025}. Optimization-based work includes hub-location and combinatorial placement methods \cite{WilleySalmon_HubSubgraph_2021,KotwiczHerniczekGerman_Placement_2024}. Operational studies further show that UAM performance is strongly affected by vertiport processes, dispatch policies, and fleet scheduling \cite{Vascik_CapacityEnvelopes_2019,Kim_RecedingHorizon_2020}.

Despite this progress, two gaps remain. First, many siting methods generate vertiport locations without evaluating them in an event-driven on-demand simulation that captures simultaneous vehicle operations and deadhead repositioning. Second, travel-time savings is often assessed using assumed non-flight penalties rather than linking those penalties to operational metrics obtained from fleet simulation.

This paper addresses these gaps through a demand--siting--simulation pipeline for on-demand UAM network design. We (i) construct spatial demand from public commuter data and transportation-hub passenger activity, (ii) generate candidate vertiport networks using clustering and screen them with range and minimum-spacing constraints, and (iii) evaluate each design using a discrete-event simulation with multi-vehicle dispatch and deadhead relocation. The simulator is coupled with an eVTOL performance model to compute completion time, service regularity, deadhead burden, and battery swaps. These outputs are then linked to a door-to-door travel-time savings analysis that estimates the allowable non-flight time budget as a function of range, ground speed, and savings target. A Greater Los Angeles case study demonstrates the proposed framework and the resulting station--fleet tradeoffs

\section{PROBLEM DEFINITION AND EVALUATION FRAMEWORK}
\label{sec:problem_definition}

\begin{figure}[b]
    \centering
    \includegraphics[width=.99\columnwidth]{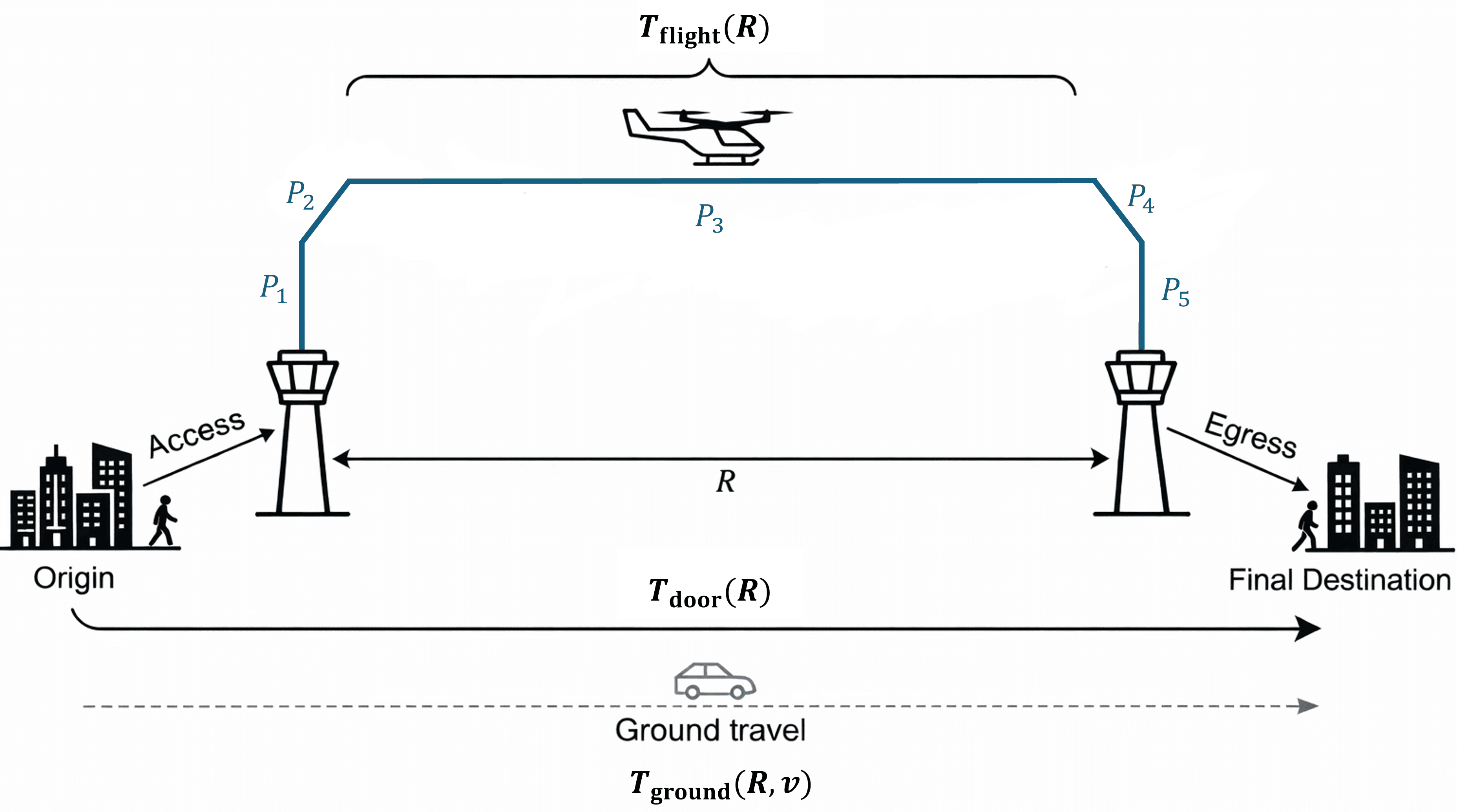}
    \caption{Door-to-door UAM trip schematic and ground alternative}
    \label{fig:UAMGround}
\end{figure}

We consider an on-demand UAM service over a geographic region with $N$ trip-ends (demand points) collected in $X=\{\mathbf{x}_n\}_{n=1}^{N}\subset\mathbb{R}^2$, where each passenger trip contributes an origin and a destination point. The system is represented by $K$ vertiports $V=\{\mathbf{v}_k\}_{k=1}^{K}\subset\mathbb{R}^2$ and a fleet of $F$ identical eVTOLs providing station-to-station flights. Each trip is modeled door-to-door with three legs: (i) ground access to an origin station, (ii) an inter-station flight, and (iii) ground egress to the final destination (Fig.~\ref{fig:UAMGround}). Let $d(\cdot,\cdot)$ denote Euclidean distance; all computations use a planar projected Coordinate Reference System (CRS), here Universal Transverse Mercator (UTM), with WGS84 used only for visualization. Passenger-to-station assignment follows a nearest-station rule, $a(\mathbf{x}_n)=\arg\min_{k\in\{1,\ldots,K\}} d(\mathbf{x}_n,\mathbf{v}_k)$, so a trip with origin $O$ and destination $D$ is routed through $(a(O),a(D))$. Aggregating assigned trip-ends yields an integer station-level OD matrix $\mathbf{Q}=[q_{ij}]$, where $q_{ij}$ is the number of passengers requesting service from station $i$ to station $j$ over the evaluation horizon.

The network design problem selects the station count $K$, station locations $V=\{\mathbf{v}_k\}_{k=1}^{K}$, and fleet size $F$. Given the study region, demand points and the induced station-level OD demand, an eVTOL performance module (per-leg flight time and energy), and fixed processing-time constants, candidate station sets are generated to ensure operational feasibility. Specifically, designs must satisfy range and spacing constraints: for all $i\neq j$, $\Delta_{\min}\le d(\mathbf{v}_i,\mathbf{v}_j)\le \Delta_{\max}$, where $\Delta_{\max}$ enforces feasible inter-station missions and $\Delta_{\min}$ prevents redundant station co-location and supports practical coverage.

To focus on network design and operations, we make a few simplifying assumptions: (i) we do not model airspace conflicts or vertiport pad-capacity limits, (ii) we assume every station can charge or swap batteries as needed, and (iii) we represent demand using the OD matrix rather than a detailed passenger arrival and queuing process. Given a candidate network design $(K,V,F)$, together with $\mathbf{Q}$, the eVTOL performance model, and fixed processing-time (e.g., boarding, deboarding, airspace clearance, and battery swap duration), a discrete-event simulation (DES) evaluates operational performance. The simulator tracks each vehicle on its own availability timeline, which allows multiple vehicles to fly simultaneously in the modeled system. For every served leg, the eVTOL performance module provides the corresponding flight time and energy consumption. The simulation then computes the principal performance measures, including mean vehicle arrival interval (MVAI) as a station-level service-regularity proxy in Equation~\eqref{eq:mvai}, completion/evacuation time $\mathrm{ET}$ using Equation~\eqref{eq:ET}, deadhead ratio (ratio of flights that carry no passenger) computed as in Equation~\eqref{eq:deadhead_ratio}, and battery swaps $\mathrm{CB}$ as in Equation~\eqref{eq:NCB}. The metrics are then aggregated in a cost function describe in Equation~\eqref{eq:cost_function} to identify the preferred design $(K^*,V^*,F^*)$ under the specified scenario assumptions.

\begin{figure}[H]
    \centering
    \includegraphics[width=\columnwidth]{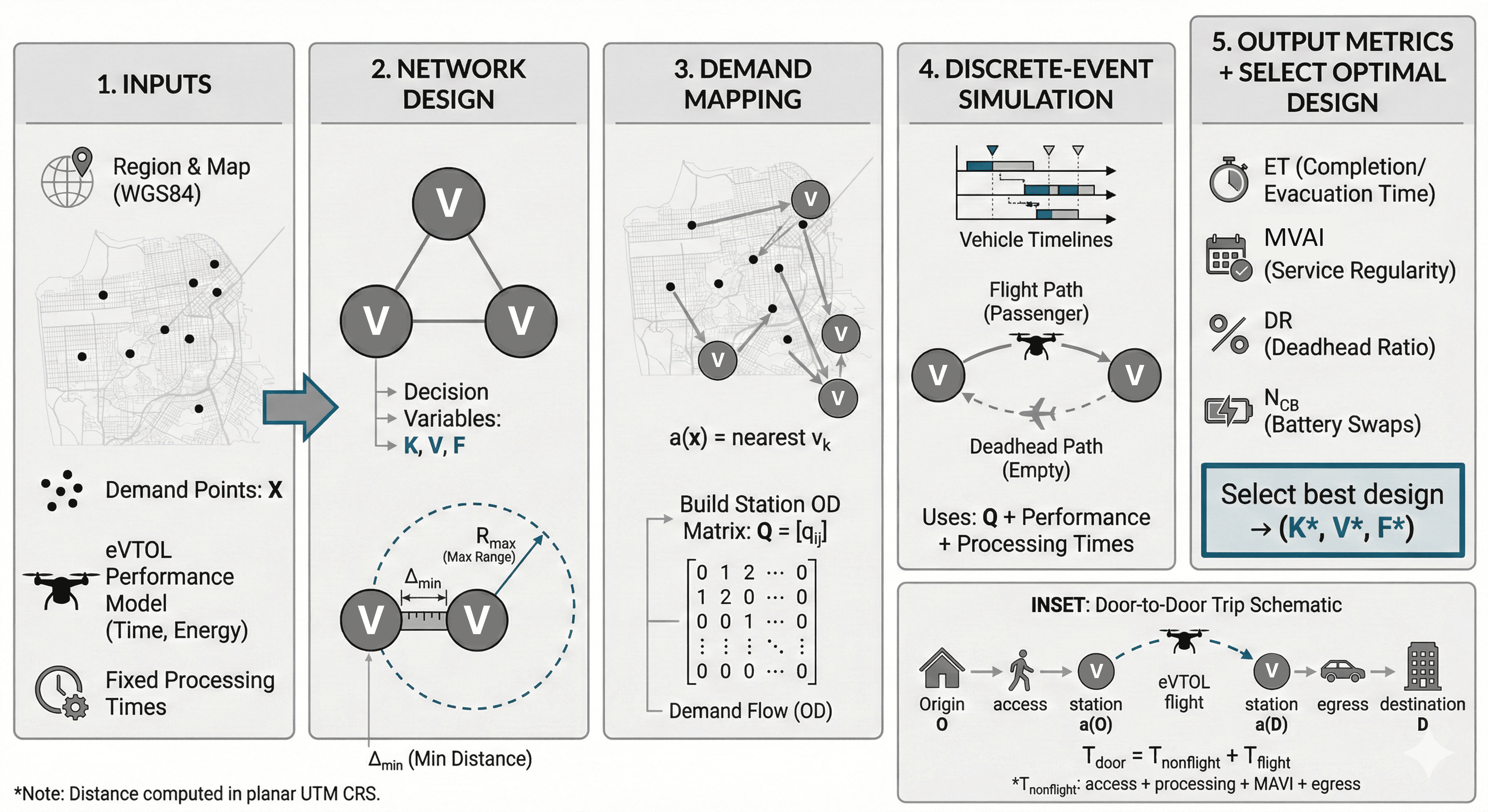}
    \caption{UAM network design and evaluation pipeline}
    \label{fig:pipeline}
\end{figure}

The door-to-door travel time for a representative trip is decomposed into flight and non-flight components using Equation~\eqref{eq:door_time}. The non-flight component aggregates access, fixed processing times, a station-level MVAI, and egress. This decomposition directly supports the travel-time savings feasibility analysis in Section~\ref{sec:tts}, which quantifies allowable non-flight time budgets under targeted time-savings requirements. The proposed pipeline is illustrated in Fig.~\ref{fig:pipeline}.

\section{METHODOLOGY}
\label{sec:methodology}
\subsection{Demand Construction and Station Siting}

UAM is an emerging transportation system, and historical usage data are not yet available to estimate demand using conventional mode-choice models (e.g., logit). Consequently, we estimate the potential demand using commuters' characteristics, including trip length, their preferred mode, and income level. Longitudinal Employer-Household Dynamics (LEHD) Origin-Destination Employment Statistics (LODES) flows \cite{census_lodes_2019} are used to compute the number of home--work commuter trips with Euclidean distance of $d \ge \Delta_{\min}$ and the corresponding trip count for home ZIP Code Tabulation Areas (ZCTA) $z$ is denoted by $C^{\text{LODES}}_{Z,z}$. American Community Survey (ACS) total commuter trips \cite{census_acs_2019_5yr} are used to reduce bias in LODES by applying a calibration factor $MF_z = TC^{\text{ACS}}_{Z,z} / TC^{\text{LODES}}_{Z,z}$ for each ZCTA, where $TC^{\text{ACS}}_{Z,z}$ and $TC^{\text{LODES}}_{Z,z}$ are denoted ACS and LODES total commuter trips, respectively. Moreover, airport and inter-city rail passenger demand component (APT) is included in the model. Denoting the total daily APT users as $T^{\text{APT}}$, ZCTA population as $P_z$, and total population of the region as $P_T=\sum_{z\in Z} P_z$, we estimate the number of daily APT in each ZCTA using $ATZ_z = T^{\text{APT}} P_z / P_T$. 

An eligibility filter $F_z \in [0,1]$ represents a near-term early-adopter demand scenario, targeting car users with annual income above \$75k \cite{census_acs_2019_5yr}, with $F_z = \mathrm{CDI75Z}_z / TC^{\text{ACS}}_{Z,z}$. This filter is not a normative equity objective or a recommendation to prioritize higher-income areas; it is only one plausible early-market demand layer. Equity-aware siting would require additional constraints or alternative demand scenarios that explicitly consider underserved communities, multimodal access, and affordability. The resulting ZCTA-level potential trip-end demand $TDZ_z$ combines outgoing and incoming trip-ends (incoming defined by destination $z$ and the same criteria) as defined in Equation~\eqref{eq:demand_modeling}. Computing demand points in ZCTA level results in many points accumulating at the representative point of each ZCTA. To obtain unique point inputs for station siting formulation, we generate the expanded demand-point set
$X_z=\{\mathbf{x}_{z_n}\}_{n=1}^{D_z}\subset\mathbb{R}^{2}$ where $D_z$ denotes the total demand in ZCTA $z$ with $N=\sum_z D_z$ and subsequently $X = \bigcup_{z \in Z} X_z \,$. For this purpose, one point is placed at each ZCTA representative point when $D_z\ge 1$, and additional points are placed uniquely by annular jittering around the representative point with acceptance constrained to the ZCTA polygon.
\begin{equation}
\label{eq:demand_modeling}
\begin{aligned}
TDZ_{z} &=
\left(C^{\mathrm{LODES}}_{Z,z}\,MF_{z} + ATZ_{z}\right)F_{z} \\
&\quad + \sum_{\substack{
z' \in Z,\; z' \neq z\\
d_{z'z}\ge \Delta_{\min},\; \operatorname{dest}(z'\!\to\! z)
}}
\left(C^{\mathrm{LODES}}_{Z,z'}\,MF_{z'} + ATZ_{z'}\right)F_{z'}
\end{aligned}
\end{equation}

We determine candidate vertiport locations by clustering the demand-point set $X$ using the $K$-means algorithm \cite{MacQueen1967} and treating the centroid of the clusters as vertiport locations. For a given $K$, $K$-means partitions $X$ into $K$ disjoint clusters $\{C_k\}_{k=1}^{K}$ and computes centroids $\{\mathbf{v}_k\}_{k=1}^{K}$ by minimizing the within-cluster sum of squares using Equation~\eqref{eq:kmeans}. This algorithm is simple and computationally efficient for large-scale demand sets, providing interpretable centroid-based station locations, and a practical baseline that has been widely adopted in prior vertiport siting studies \cite{Jeong_Kmeans_Vertiports_2021,Guo_ClusteringComparison_2025}.
\begin{equation}
\min_{\{C_k\},\,\{\mathbf{v}_k\}}
\sum_{k=1}^{K}\sum_{\mathbf{x}_n\in C_k}\left\lVert \mathbf{x}_n-\mathbf{v}_k \right\rVert_2^2
\label{eq:kmeans}
\end{equation}

 To select an appropriate station count $K$, we evaluate clustering quality using the silhouette coefficient \cite{Rousseeuw1987Silhouettes}. For each demand point $\mathbf{x}_k$, let $m(k)$ denote the mean distance to points in its assigned cluster and $m\prime(k)$ the minimum mean distance to points in any other cluster. The silhouette value is $ss(k)=\frac{m\prime(k)-m(k)}{\max\{m(k),\,m\prime(k)\}} \in [-1,1]$. We used the average silhouette over all points to compare the candidate values of $K$. Candidate station sets are retained only if the resulting inter-station distances satisfy the mission-feasibility constraints in Section~\ref{sec:problem_definition}, ensuring that the siting output can be passed directly to the network simulation stage.
\subsection{Discrete-Event On-Demand Operations Model}
\label{subsec:des_ops}
Given a candidate network design $(K,V,F)$ and its associated station-level OD demand matrix $\mathbf{Q}$ defined in Section~\ref{sec:problem_definition}, we evaluate on-demand operations using a DES. The simulator keeps track of (i) how many trips are still left to serve in $\mathbf{Q}$ and (ii) when and where each vehicle becomes available again after finishing its current leg. Each event happens when one of the eVTOLs becomes available to take its next trip. At that event, the vehicle either serves a passenger-carrying OD request or relocates via a deadhead leg when local demand is exhausted. The simulation terminates when all OD entries have been served.

\textit{Monte Carlo demand and initial conditions.}
To capture day-to-day variability in route-level demand, we optionally evaluate designs under Monte Carlo runs in which OD flows are stochastically redistributed while preserving a fixed total passenger budget per run. This randomized redistribution is used as an operational stress test rather than as a calibrated empirical OD forecast. It allows each candidate design to be evaluated under multiple route-demand realizations with the same total demand level, so the comparison focuses on how station--fleet configurations respond to demand imbalance and randomized initial vehicle placement. For run $o\in\{1,\dots,O\}$, we generate an integer OD matrix $\mathbf{Q}^{(o)}=\big[q_{ij}^{(o)}\big],q_{ij}^{(o)}\in\mathbb{Z}_{\ge 0}, q_{ii}^{(o)}=0, \sum_{i\neq j} q_{ij}^{(o)}=D$, where $D$ is the total number of passengers trips in each run. Off-diagonal entries are sampled sequentially using a remaining-demand budget $D_{\mathrm{rem}}$ (initialized as $D$) through the following steps: (i) iterating over OD pairs $(i,j)$ with $i\neq j$, we draw $\tilde q_{ij}\sim U\left(0, D_{\mathrm{rem}}/K\right)$, (ii) set $q_{ij}=\mathrm{round}(\tilde q_{ij})$, and update $D_{\mathrm{rem}}\leftarrow D_{\mathrm{rem}}-q_{ij}$, and finally (iii) any residual $D_{\mathrm{rem}}$ is assigned to the final OD pair in the iteration order to enforce total of $D$. In each run, initial eVTOL locations are randomized by assigning each vehicle to an initial station drawn uniformly from $\{1,\dots,K\}$, which avoids biasing early service toward any particular station configuration.

\textit{Event logic: dispatch and deadhead relocation.}
At each event, the next-available vehicle is selected, and its current station is denoted by $s$. If station $s$ has remaining outbound demand (i.e., $\sum_{j\neq s} q_{sj}>0$), the vehicle is dispatched on a passenger-carrying leg from $s$ to a destination station selected from the set of feasible OD requests (the implementation prioritizes the largest remaining OD entry, with ties broken by choosing the nearest destination). If station $s$ has no remaining outbound demand but demand remains elsewhere in the network, the vehicle performs a deadhead relocation. Consistent with the implemented simulation rule, relocation uses an admissibility constraint to prevent multiple vehicles from repeatedly concentrating at the same demand-rich station. Define the set of stations with remaining outbound demand and the set of currently occupied stations as $S_{\mathrm{dem}}=\left\{k:\sum_{j\neq k} q_{kj}>0\right\},  
S_{\mathrm{occ}}=\left\{k:\text{at least one vehicle is currently at station }k\right\}$. Candidate relocation destinations are restricted to $S_{\mathrm{cand}}=S_{\mathrm{dem}}\setminus S_{\mathrm{occ}}$, and the deadhead destination is chosen as the nearest admissible station as in $k^{\star}=\arg\min_{k\in S_{\mathrm{cand}}} d(\mathbf{v}_{s},\mathbf{v}_{k})$. By sending idle eVTOLs to the nearest station with unmet demand, deadhead flights help eVTOLs re-enter the service network and can speed up the overall completion process.

\textit{Per-leg flight simulation and minimum-energy altitude.}
Each flown leg (passenger-carrying or deadhead) between stations $i$ and $j$ is evaluated by a point-mass eVTOL model implemented in MATLAB/Simulink (R2025). The mission profile follows five phases of vertical climb/VC ($p_1$), climb ($p_2$), cruise ($p_3$), descent ($p_4$), and vertical descent/VD ($p_5$) as shown in Fig.~\ref{fig:UAMGround}. For a given inter-station distance $R_{ij}=d(\mathbf{v}_{i},\mathbf{v}_{j})$, cruise altitude is selected to minimize energy consumption for that leg. Denoting total leg energy by $E_l(R_{ij},h)$ and leg flight time by $T_{\mathrm{flight}}(R_{ij},h)$ at cruise altitude $h$, the model uses $h^{\star}(R_{ij})=\arg\min_{h} E_l(R_{ij},h)$, $T_{\mathrm{flight}}(R_{ij})=T_{\mathrm{flight}}(R_{ij},h^{\star}(R_{ij}))$, and $E_{l_{ij}}=E_{l}(R_{ij},h^{\star}(R_{ij}))$. Within each phase, the point-mass kinematics and energy consumption are simulated with dynamics described concisely in Equations~\eqref{eq:point_mass1}-\eqref{eq:point_mass6} with $P= \{p_1,p_2,\mathrm{or}\ p_3\}$. Interested readers can refer to \cite{Saarlas2006AircraftPerformance} and \cite{SeddonNewman2011BasicHelicopterAerodynamics} for details on point-mass and energy-consumption modeling. The trajectory is controlled using a velocity-pursuit guidance law that aligns the velocity vector with the line-of-sight (LOS) to the destination. Let $\mathbf{r}=[X,Y,h]^\top$ be the eVTOL position, $\mathbf{r}_{\mathrm{dest}}=[X_{\mathrm{dest}},Y_{\mathrm{dest}},h_{\mathrm{dest}}]^\top$ be the destination position, $\hat{\mathbf{e}}_{\mathrm{LOS}}=(\mathbf{r}_{\mathrm{dest}}-\mathbf{r})/\|\mathbf{r}_{\mathrm{dest}}-\mathbf{r}\|$, and $\hat{\mathbf{e}}_{v_P}=\mathbf{v_P}/\|\mathbf{v_P}\|$ be the unit velocity direction with $\mathbf{v}_{P}=[\dot{X}_{P},\dot{Y}_{P},\dot{h}_{P}]^\top$ and $V_p=\|\mathbf{v_P}\|$. The commanded lateral acceleration and yaw angle rate are computed as in Equation~\eqref{eq:guidance}, which regulates heading toward the destination. Finally, each trip-leg flight time and energy consumption is computed by Equation~\eqref{eq:time_energy_leg}. Description of the parameters in the flight equations and their values are displayed in Table~\ref{tab:params}.
\begin{equation}
\label{eq:point_mass1}
\dot{X_{P}} = V_{p}\cos\gamma_{P}\cos\psi_{P},\quad
\dot{Y_{P}} = V_{p}\cos\gamma_{P}\sin\psi_{P},\quad
\dot{h_{P}} = V_{p}\sin\gamma_{P}
\end{equation}
\begin{equation}
\label{eq:point_mass2}
\dot{\gamma}_{P} = \frac{1}{m V_{P}}\left(L_{P} + T_{P}\sin(\epsilon) - W\cos(\gamma_{P})\right),~L_{P} = \frac{1}{2}\rho V_{P}^2S_\mathrm{wing}C_{L_P}
\end{equation}
\begin{equation}
\label{eq:point_mass3}
\dot{V}_{P} = \frac{1}{m} \left(T_{P}\cos(\epsilon) - D_{P} - W\sin(\gamma_{P})\right),~ D_{P} = \frac{1}{2}\rho V_{P}^2S_\mathrm{wing}C_{D_P}
\end{equation}
\begin{equation}
\label{eq:point_mass4}
C_{L_P}=C_{L_0}+C_{L_\alpha}\alpha_{P}\quad C_{D_P}=C_{D_0}+C_{D_{C_{L}^2}}C_{L_P}^2+C_{D_{C_L}}C_{L_P}
\end{equation}
\begin{equation}
\label{eq:point_mass5}
T_P = D_P+W\sin\gamma_P \quad E_{P} = \int \frac{T_{P}\,V_{P}}{\eta_{\mathrm{prop}}}\, dt\quad t_{p_1/p_5}=\frac{h_{p_1/p_5}}{V_{p_1/p_5}}
\end{equation}
\begin{equation}
\label{eq:point_mass6}
\scalebox{1}{$
E_{p_1/p_5}=\frac{W h_{p_1/p_5}}{\eta_{\mathrm{rotor}}V_{p_1/p_5}}
\Bigg(
\frac{V_{p_1/p_5}}{2}
+ \sqrt{\frac{2W}{2\rho \pi n_{p_1/p_5}D_{\text{rotor}}^2}
      + \frac{V_{p_1/p_5}^{2}}{4}}
\Bigg)
$}
\end{equation}
\begin{equation}
\label{eq:guidance}
\mathbf{a}_{c_y,P}=\|\mathbf{v_P}\|\left(\hat{\mathbf{e}}_{v_{P}}\times \hat{\mathbf{e}}_{\mathrm{LOS}}\right)\times \hat{\mathbf{e}}_{v_{P}},\quad
\dot{\psi_{P}}
=
\frac{\left\lVert \mathbf{a}_{c_y,P} \right\rVert}{\left\lVert \mathbf{v}_{P} \right\rVert}
\end{equation}
\begin{equation}
\label{eq:time_energy_leg}
T_{\mathrm{flight}}=\sum_{p=p_{1}}^{p_{5}}t_p,\quad E_{l}=\sum_{p=p_{1}}^{p_{5}}E_p
\end{equation}

\textit{Mean Vehicle Arrival Interval (MVAI) metric}
In addition to completion time, the simulator reports a station-level service-regularity proxy derived from vehicle arrival timestamps. Let $\{t_{s,1},t_{s,2},\dots,t_{s,B_s}\}$ denote the time-ordered sequence of vehicle arrivals recorded at station $s$ over the simulation horizon, where $B_s$ is the number of recorded arrivals at station $s$. The station-level MVAI is calculated by Equation~\eqref{eq:mvai}.
\begin{equation}
\label{eq:mvai}
\Delta_s=\frac{1}{B_s-1}\sum_{b=2}^{B_s}\left(t_{s,b}-t_{s,b-1}\right),\qquad
\mathrm{MVAI}=\frac{1}{K}\sum_{s=1}^{K}\Delta_s
\end{equation}

As implemented, $\mathrm{MVAI}$ is a station-level vehicle-arrival regularity metric, not a passenger waiting-time estimate. Since the DES does not model passenger arrivals or queues, lower $\mathrm{MVAI}$ indicates more frequent vehicle availability.

\textit{Network evacuation/completion time and battery swaps.}
Let $T_{\mathrm{cum_f}}$ denote the cumulative flight time of vehicle $f$ by the end of the simulation, and let $T_{\max}=\max(T_{\mathrm{cum_f}})$ be the overall maximum flight time over the fleet. The reported evacuation/completion time is the sum of $T_{\max}$  and fixed processing time in each Monte Carlo run as defined by Equation~\eqref{eq:ET}, where $\mathrm{CB}$ is the number of battery-change events, $T_{CB}$ is the battery-change duration, $N_{P}$ is the number of passenger-carrying flights, $T_{\mathrm{board}}$ and $T_{\mathrm{deboard}}$ are boarding and deboarding times (counted per served flight), $N_{\mathrm{tot}}$ is the total number of flown legs (including deadheads), and $T_{\mathrm{clr}}$ is a fixed airspace clearance time per flight.
\begin{equation}
\label{eq:ET}
\mathrm{ET} = T_{\max} + \mathrm{CB}T_{CB} + N_{P}\left(T_{\mathrm{board}}+T_{\mathrm{deboard}}\right) + N_{\mathrm{tot}}T_{\mathrm{clr}}
\end{equation}

Battery swaps are computed from total network energy consumption $E_{\mathrm{net}}=\sum_{l\in L}E_{l}$ accumulated over all flown legs $L$ and a fixed per-battery available energy $E_{\mathrm{bat}}$ as in Equation~\eqref{eq:NCB}.
\begin{equation}
\label{eq:NCB}
\mathrm{CB}=\mathrm{ceil}\!\left(\frac{E_{\mathrm{net}}}{E_{\mathrm{bat}}}\right)
\end{equation}

Deadhead burden is summarized by the deadhead ratio, where $N_{\mathrm{dead}}$ is the number of deadhead legs executed during the simulation.
\begin{equation}
\label{eq:deadhead_ratio}
\mathrm{DHR}=\frac{N_{\mathrm{dead}}}{N_{\mathrm{tot}}}
\end{equation}

\begin{figure}[H]
    \centering
    \includegraphics[width=\columnwidth]{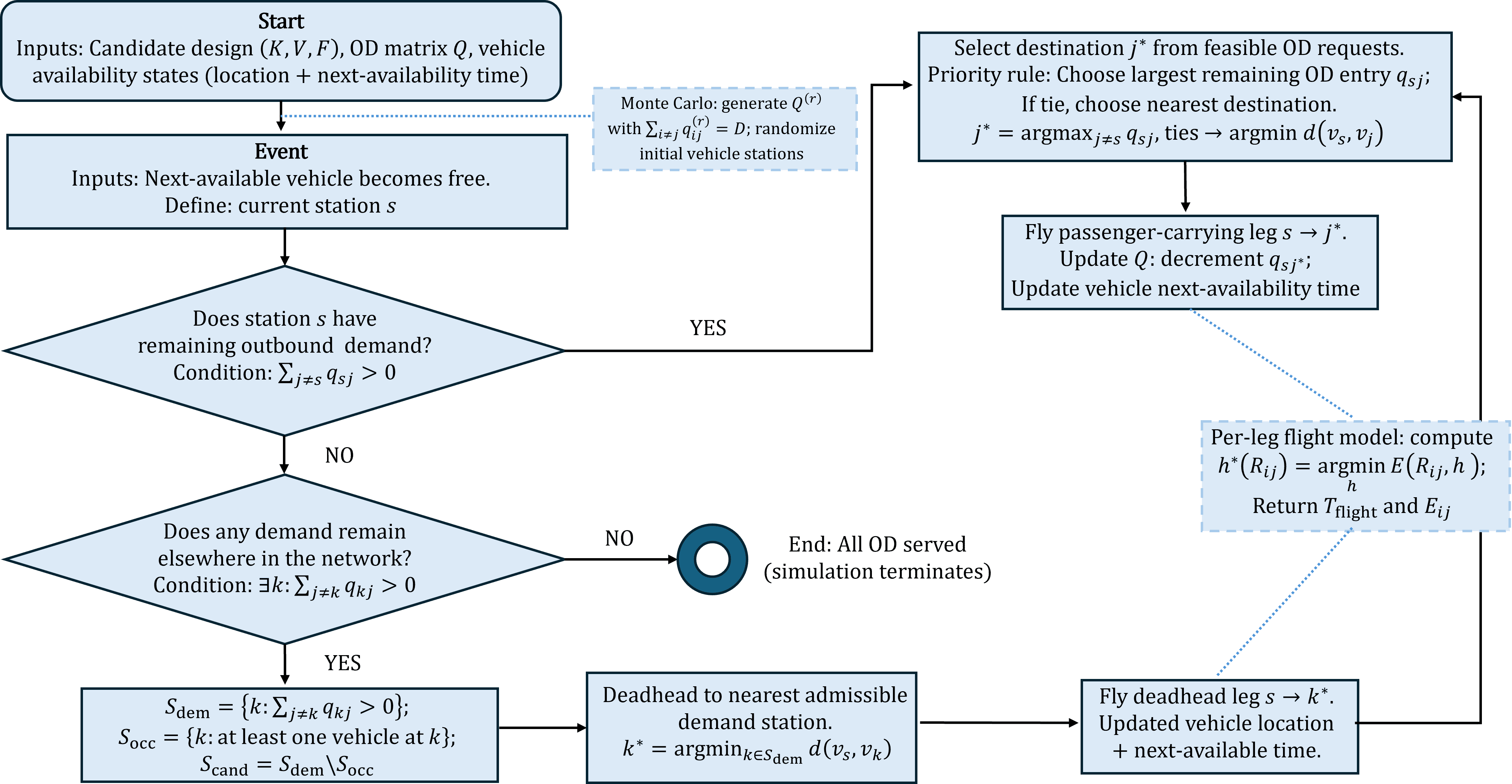}
    \caption{DES Dispatch and Relocation Logic}
    \label{fig:DSE}
\end{figure}

The aforementioned metrics $(ET,\mathrm{MVAI},\mathrm{DHR},\mathrm{CB})$ as well as Number of Stations and Number of eVTOLs, denoted by $\mathrm{NS}$ and $\mathrm{NE}$, respectively, are then passed to the design selection criterion introduced in Section~\ref{sec:design_selection} to identify preferred network configurations under the stated assumptions. Fig.~\ref{fig:DSE} summarizes DES dispatch and relocation logic explained in this section.

\subsection{Design Selection Criterion}
\label{sec:design_selection}
For each candidate network design $(K,V,F)$ generated in this section and evaluated via the DES, we compute the operational performance metrics $\mathrm{MVAI}$, $\mathrm{ET}$, $\mathrm{CB}$, and $\mathrm{DHR}$. When Monte Carlo runs are enabled, each metric is first aggregated over runs (e.g., by the sample mean) to obtain a single performance estimate per candidate design, denoted generically by $\overline{\mathrm{ET}}(K,V,F)$, $\overline{\mathrm{MVAI}}(K,V,F)$, $\overline{\mathrm{DHR}}(K,V,F)$, and $\bar{N}_{CB}(K,V,F)$. To enable a weighted tradeoff across heterogeneous metrics, we normalize each aggregated metric over the candidate design set $C$ using min--max scaling described in Equation~\eqref{eq:norm_metric}, where $\bar{M}$ denotes the aggregated value of a metric (e.g., $\overline{\mathrm{ET}}$) and $\widehat{M}\in[0,1]$ is its normalized counterpart. We then define the weighted cost function in Equation~\eqref{eq:cost_function}, with nonnegative weights satisfying $w_{\mathrm{ET}} + w_{\mathrm{MVAI}} + w_{\mathrm{DHR}} + w_{\mathrm{CB}} + w_{\mathrm{NE}} + w_{\mathrm{NS}} = 1$ and $w_i > 0$. We set $w_{\mathrm{ET}} = w_{\mathrm{MVAI}} = 2w_{\mathrm{CB}} = 4w_{\mathrm{DHR}} = 4w_{\mathrm{NE}} = 4w_{\mathrm{NS}}$, which prioritizes temporal metrics, followed by the energy-related term, and finally the remaining cost-related terms. These weights are not intended to represent a universal economic cost model; rather, they encode the planning preference adopted in this study, where reducing completion time and improving service regularity are prioritized over secondary operational and infrastructure penalties. The preferred design is selected as $(K^{\ast},V^{\ast},F^{\ast})=\arg\min_{(K,V,F)\in {C}} J(K,V,F)$.

\begin{equation}
\label{eq:norm_metric}
\widehat{M}(K,V,F)=\frac{\bar{M}(K,V,F)-\min_{(K,V,F)\in {C}}\bar{M}}
{\max_{(K,V,F)\in {C}}\bar{M}-\min_{(K,V,F)\in {C}}\bar{M}}
\end{equation}

\begin{equation}
\label{eq:cost_function}
\begin{split}
J(K,V,F) &=
w_{\mathrm{ET}}\,\widehat{\mathrm{ET}}
+ w_{\mathrm{MVAI}}\,\widehat{\mathrm{MVAI}}
+ w_{\mathrm{DHR}}\,\widehat{\mathrm{DHR}} \\
&\quad
+ w_{\mathrm{CB}}\,\widehat{\mathrm{CB}}
+ w_{\mathrm{NE}}\,\widehat{NE}
+ w_{\mathrm{NS}}\,\widehat{NS}
\end{split}
\end{equation}

\subsection{Door-to-Door Travel-Time Savings Feasibility}
\label{sec:tts}

Following the door-to-door formulation introduced in Section~\ref{sec:problem_definition}, we express the travel time of a representative trip as the sum of flight and non-flight components as in Equation~\eqref{eq:door_time}. Here $R$ denotes the associated station-to-station separation (e.g., $R=d(\mathbf{v}_i,\mathbf{v}_j)$ for some $i\neq j$).
\begin{equation}
\label{eq:door_time}
T_{\mathrm{door}}(R)=T_{\mathrm{nonflight}}(R)+T_{\mathrm{flight}}(R)
\end{equation}

To evaluate whether UAM can achieve a target fractional travel-time savings relative to a ground alternative, we model the corresponding ground travel time using an effective average ground speed $v$, routing (circuity) factor $\kappa = 1.42$ \cite{UberElevate2016FastForwarding}, and access/egress radius of $\delta \leq5$km as in $T_{\mathrm{ground}}(R;v)=\kappa\,\frac{R+\delta}{v}$. For a desired savings target $s\in(0,1)$, feasibility requires the UAM door-to-door time to be no greater than a $(1-s)$ fraction of the ground travel time or $T_{\mathrm{door}}(R)\le (1-s)\,T_{\mathrm{ground}}(R;v)$.
Substituting Equation~\eqref{eq:door_time} into this results in an allowable non-flight time budget (slack) for each $(R,v,s)$ of $\tau(R;v,s)=(1-s)\,T_{\mathrm{ground}}(R;v)-T_{\mathrm{flight}}(R)$, and saving target is achievable only if $T_{\mathrm{nonflight}}(R)\le \tau(R;v,s)$(Fig.~\ref{fig:UAMGround}). A positive slack $\tau>0$ indicates that a non-flight time budget exists to meet the desired savings, whereas $\tau<0$ implies that the target savings cannot be met even with zero non-flight time.

\section{CASE STUDY SETUP}
\label{sec:case_study}
We demonstrate the proposed demand--siting--simulation pipeline (Fig.~\ref{fig:pipeline}) on the Greater Los Angeles region. All spatial computations are performed in the UTM projected CRS using the Euclidean metric $d(\cdot,\cdot)$, while WGS84 coordinates are used only for map visualization (Section~\ref{sec:problem_definition}). Demand is constructed using publicly available commuter datasets (Section~\ref{sec:methodology}) together with an airport and inter-city rail passenger component, yielding a set of trip-end demand points $X=\bigcup_{z\in Z}X_z$ and the station-level OD matrix $\mathbf{Q}$. For this region and the adopted early-user eligibility filter, the resulting candidate passenger pool is on the order of $2.96\times 10^{5}$ trip-ends (Section~\ref{sec:methodology}), which serves as the baseline demand from which scenario volumes are defined below.

\begin{figure}[tb]
    \centering
    \includegraphics[width=.99\columnwidth]{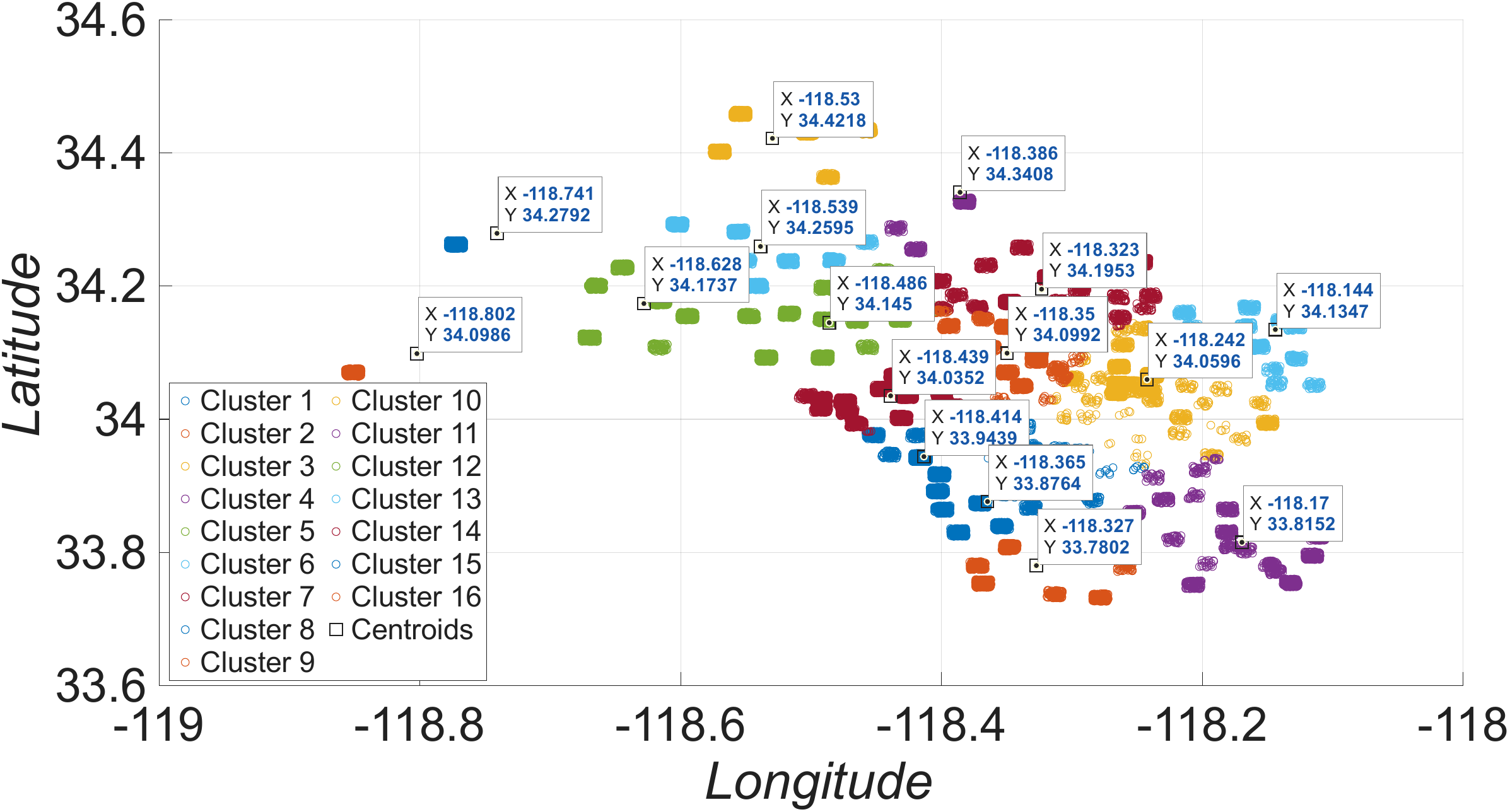}
    \caption{Centroids of the 16-clusters case used as candidate station locations; station latitudes and longitudes are annotated}
    \label{fig:kmenas}
\end{figure}

\begin{figure}[tb]
    \centering
    \includegraphics[width=.99\columnwidth]{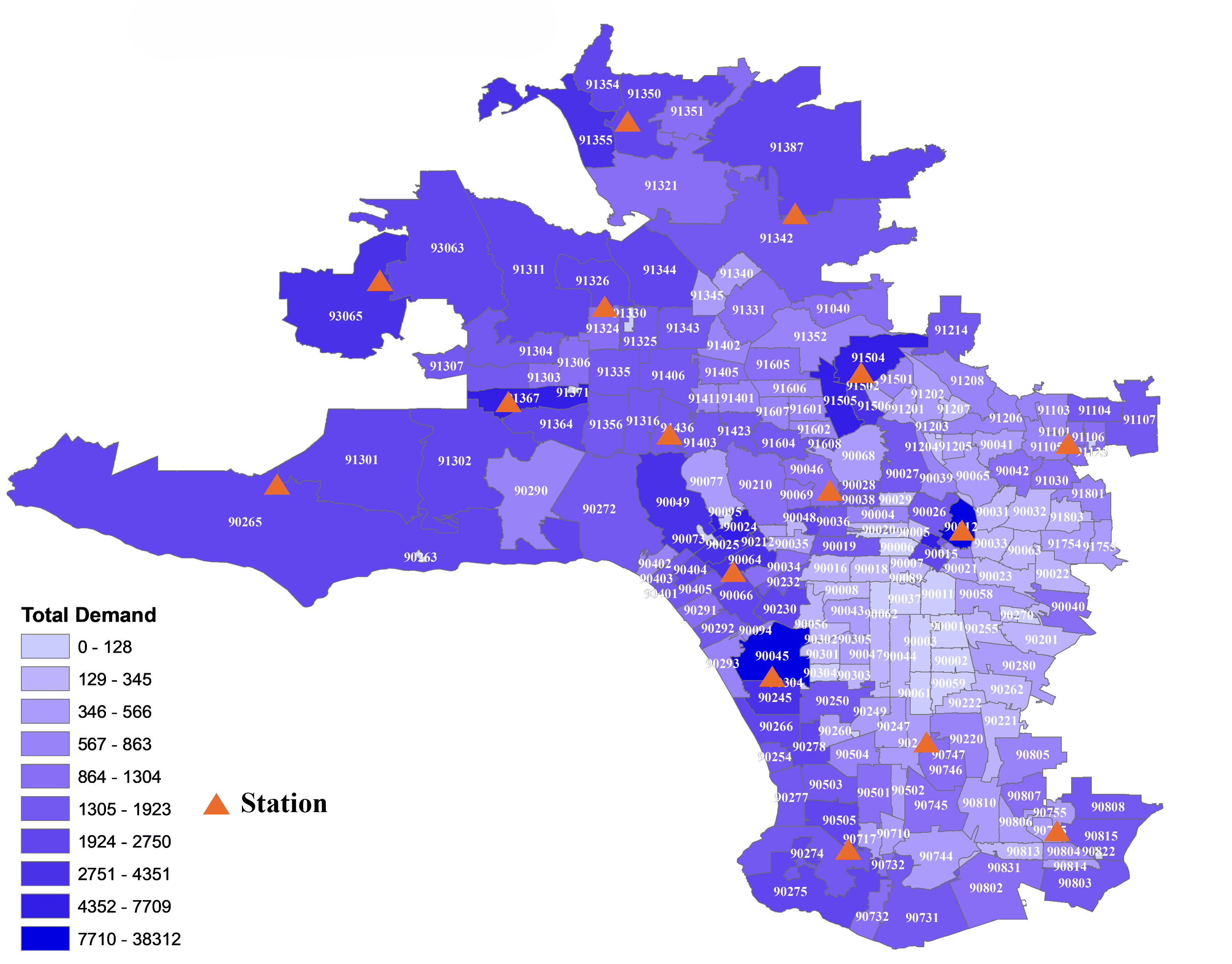}
    \caption{Stations mapping over the study area}
    \label{fig:stations_location}
\end{figure}

We evaluate a family of candidate network designs $(K,V,F)$ subject to the feasibility constraints in Section~\ref{sec:problem_definition}. Candidate station locations $V=\{\mathbf{v}_k\}_{k=1}^{K}$ are generated by the $K$-means siting procedure in Equation~\eqref{eq:kmeans} and screened to satisfy the feasibility condition of $\Delta_{\min}\le d(\mathbf{v}_i,\mathbf{v}_j)\le \Delta_{\max}$ for all $i\neq j$ (Figs.~\ref{fig:kmenas} and~\ref{fig:stations_location}). Table~\ref{tab:clustering_results} shows that the feasibility constraint caps the tested network at 16 stations. Increasing the station count beyond 16 leads to infeasible missions and a lower silhouette value. For each feasible station set ($K=\{4,8,12,16\}$), we test multiple fleet sizes $F=\{2:2:K\}$ to quantify station--fleet tradeoffs under on-demand operations. Scenario demand levels are defined as representative operational stress-test volumes ($D\in\{100,500,1000,2000\}$ passenger trips). The spatial demand model determines the demand-weighted station locations, while the Monte Carlo OD generation evaluates the operational behavior of those station sets under randomized station-level route demand with fixed total volume. Performance metrics are aggregated across Monte Carlo runs before selecting the preferred design $(K^\ast,V^\ast,F^\ast)$. Table~\ref{tab:params} summarizes the parameters and their values used in the case study.


\begin{table}[tb]
\caption{Case-study parameters and operational constants}
\label{tab:params}
\centering
\resizebox{\columnwidth}{!}{
\begin{tabular}{|c|l|c|}
\hline
\textbf{Parameter (unit)} & \textbf{Description} & \textbf{Value} \\
\hline
$\Delta_{\max}\,(\mathrm{km})$ & Max inter-station distance & $81$ \\
\hline
$\Delta_{\min}\,(\mathrm{km})$ & Min station spacing & $10$ \\
\hline
$O$ & Number of runs & $250$ \\
\hline
$E_{\mathrm{bat}}\,(\mathrm{MJ})$ & Battery usable energy & $339$ \\
\hline
$\{T_{CB},T_{\mathrm{board}},T_{\mathrm{deboard}},T_{\mathrm{clr}}\}\,(\mathrm{s})$
& Swap, board, deboard, clearance times & $\{90,240,150,45\}$ \\
\hline
$v\,(\mathrm{km/h})$ & Average ground speed & $\{50,80\}$ \\
\hline
$\{\gamma_{p_{2}},\gamma_{p_{3}},\gamma_{p_{4}}\}\,(^\circ)$ & Flight-path angles (climb, cruise, descent) & $\{5,0,-4\}$ \\
\hline
$\{m,m_{pax}\}\,(\mathrm{kg})$ & Empty takeoff, passenger mass & $\{1620,90\}$ \\
\hline
$n_{pax}$ & Number of passengers & $\{0,1,2,3,4\}$ \\
\hline
$W$ (N) & Total flight weight ($m+n_{pax}\times m_{pax} \times 9.81$) & [15892 , 19424]\\
\hline
$\{h_{p_{1}},h_{p_{5}}\}\,(\mathrm{m})$ & VC/VD altitude (MSL) & $15$ \\
\hline
$\{V_{p_{1}},V_{p_{2}},V_{p_{3}},V_{p_{4}},V_{p_{5}}\}\,(\mathrm{km/h})$
& Airspeeds (VC, climb, cruise, descent, VD) & $\{11.5,209,263,252,11.5\}$ \\
\hline
$\{n_{p_{1}/p_{5}},n_{p_{3}},n_{p_{2}/p_{4}}\}$ & Active motors (VC/VD, cruise, climb/descent) & $\{8,4,4\}$ \\
\hline
$\{\eta_{\mathrm{rotor}},\eta_{\mathrm{prop}} \}$ & Rotor and propeller efficiency & $\{0.65, 0.75 \}$ \\
\hline
$D_{\mathrm{rotor}}\,(\mathrm{m})$ & Rotor diameter & $1.78$ \\
\hline
$S_{\mathrm{wing}}\,(\mathrm{m^2})$ & Wing area & $2 \times 8.194$ \\
\hline
$\epsilon\,(^\circ)$ & Installation angle & $0$ \\
\hline
\end{tabular}}
\end{table}

\begin{table}[tb]
\caption{Clustering results}
\label{tab:clustering_results}
\centering
\scriptsize
\setlength{\tabcolsep}{3pt}
\renewcommand{\arraystretch}{1.10}
\begin{tabular}{|c|c|c|c|c|}
\hline
\textbf{\# Stations} & \textbf{Avg.\ $ss(k)$} & \textbf{Infeasible missions} &
\textbf{Min dist.\ (km)} & \textbf{Max dist.\ (km)} \\
\hline
4  & 0.6297 & 0 & 20 & 55 \\
\hline
8  & 0.6359 & 0 & 14 & 70 \\
\hline
12 & 0.6713 & 0 & 12 & 76 \\
\hline
16 & 0.6949 & 0 & 11 & 78 \\
\hline
18 & 0.6915 & 2 &  8 & 78 \\
\hline
\end{tabular}
\end{table}

\section{RESULTS}
\label{sec:results}
This section reports the Greater Los Angeles case-study results. Candidate designs $(K,V,F)$ are evaluated using the DES in Section~\ref{subsec:des_ops}, and the preferred design for each demand scenario is selected by minimizing $J$ in Equation~\eqref{eq:cost_function}. We report how $(K^\ast,F^\ast)$ scales with demand and summarize the main operational tradeoffs using $\mathrm{ET}$, $\mathrm{MVAI}$, $\mathrm{DHR}$, $\mathrm{CB}$, $\mathrm{NE}$, and $\mathrm{NS}$.

\subsection{Preferred Candidate Designs Across Demand Scenarios}
\label{subsec:results_opt}

Using the selection criterion in Equation~\eqref{eq:cost_function}, we obtain the preferred candidate design $(K^{\ast},V^{\ast},F^{\ast})$ for each demand level $D$. Table~\ref{tab:opt_networks} reports the selected station--fleet combinations and their DES performance. The station-location component $V^{\ast}$ is determined by the corresponding feasible $K$-means station set; for compactness, the table reports $(K^{\ast},F^{\ast})$, while the coordinates and spatial layout of the candidate station sets are shown in Figs.~\ref{fig:kmenas} and~\ref{fig:stations_location}. As demand increases, the preferred design expands from $(K^{\ast},F^{\ast})=(4,4)$ at $D=100$ to $(8,8)$ at $D=500$--$1000$, and to the maximum feasible station count at high demand, with $(16,12)$ at $D=2000$. This pattern indicates that both infrastructure scale and fleet size become more important as total passenger volume increases. Across scenarios, $\mathrm{DHR}$ remains relatively stable ($\approx 22\%$--$30\%$), while $\mathrm{CB}$ increases markedly with demand, consistent with higher total energy consumption.

Table~\ref{tab:extreme_2000} illustrates a fleet-abundant sensitivity case (here $D=2000$) in which $F=4K$ for several station counts. As $F$ increases, both $\mathrm{ET}$ and $\mathrm{MVAI}$ decrease substantially, indicating that temporal performance is fleet-limited at high demand. This improvement is accompanied by higher battery-swap counts ($\mathrm{CB}$), while $\mathrm{DHR}$ remains nonzero, confirming that deadhead flights are still needed to move vehicles to where demand remains and keep the network running smoothly. This case is not intended to represent a cost-optimal deployment; rather, it isolates the effect of relaxing fleet availability and shows that adding vehicles alone cannot remove repositioning needs when OD demand is spatially imbalanced.

\begin{table}[tb]
\caption{Preferred candidate design and performance across demand scenarios}
\label{tab:opt_networks}
\centering
\scriptsize
\setlength{\tabcolsep}{4pt}
\renewcommand{\arraystretch}{1.15}
\begin{tabular}{|c|c|c|c|c|c|}
\hline
\textbf{$D$} & \textbf{$(K^{\ast},F^{\ast})$} & \textbf{$\mathrm{ET}$ (h)} & \textbf{$\mathrm{MVAI}$ (min)} & \textbf{$\mathrm{DHR}$ (\%)} & \textbf{$\mathrm{CB}$} \\
\hline
100  & (4, 4)    & 2  & 18 & 22 & 15  \\
\hline
500  & (8, 8)    & 6  & 26 & 28 & 77  \\
\hline
1000 & (8, 8)    & 11 & 34 & 30 & 145 \\
\hline
2000 & (16, 12)  & 15 & 37 & 29 & 337 \\
\hline
\end{tabular}
\end{table}

\begin{table}[tb]
\caption{Fleet-abundant extreme case for $D=2000$ with $F=4K$}
\label{tab:extreme_2000}
\centering
\scriptsize
\setlength{\tabcolsep}{4pt}
\renewcommand{\arraystretch}{1.15}
\begin{tabular}{|c|c|c|c|c|c|}
\hline
\textbf{$K$} & \textbf{$F$} & \textbf{$\mathrm{ET}$ (h)} & \textbf{$\mathrm{MVAI}$ (min)} & \textbf{$\mathrm{DHR}$ (\%)} & \textbf{$\mathrm{CB}$} \\
\hline
4  & 16 & 10 & 30 & 24 & 418 \\
\hline
8  & 32 & 7  & 15 & 28 & 400 \\
\hline
12 & 48 & 4  & 13 & 23 & 490 \\
\hline
16 & 64 & 3  & 10 & 24 & 520 \\
\hline
\end{tabular}
\end{table}

\subsection{Station--Fleet Tradeoffs and Objective Landscape}
\label{subsec:results_tradeoff}
\begin{figure*}[!t]
    \centering
    \begin{subfigure}[t]{0.45\textwidth}
        \centering
        \includegraphics[width=0.94\linewidth]{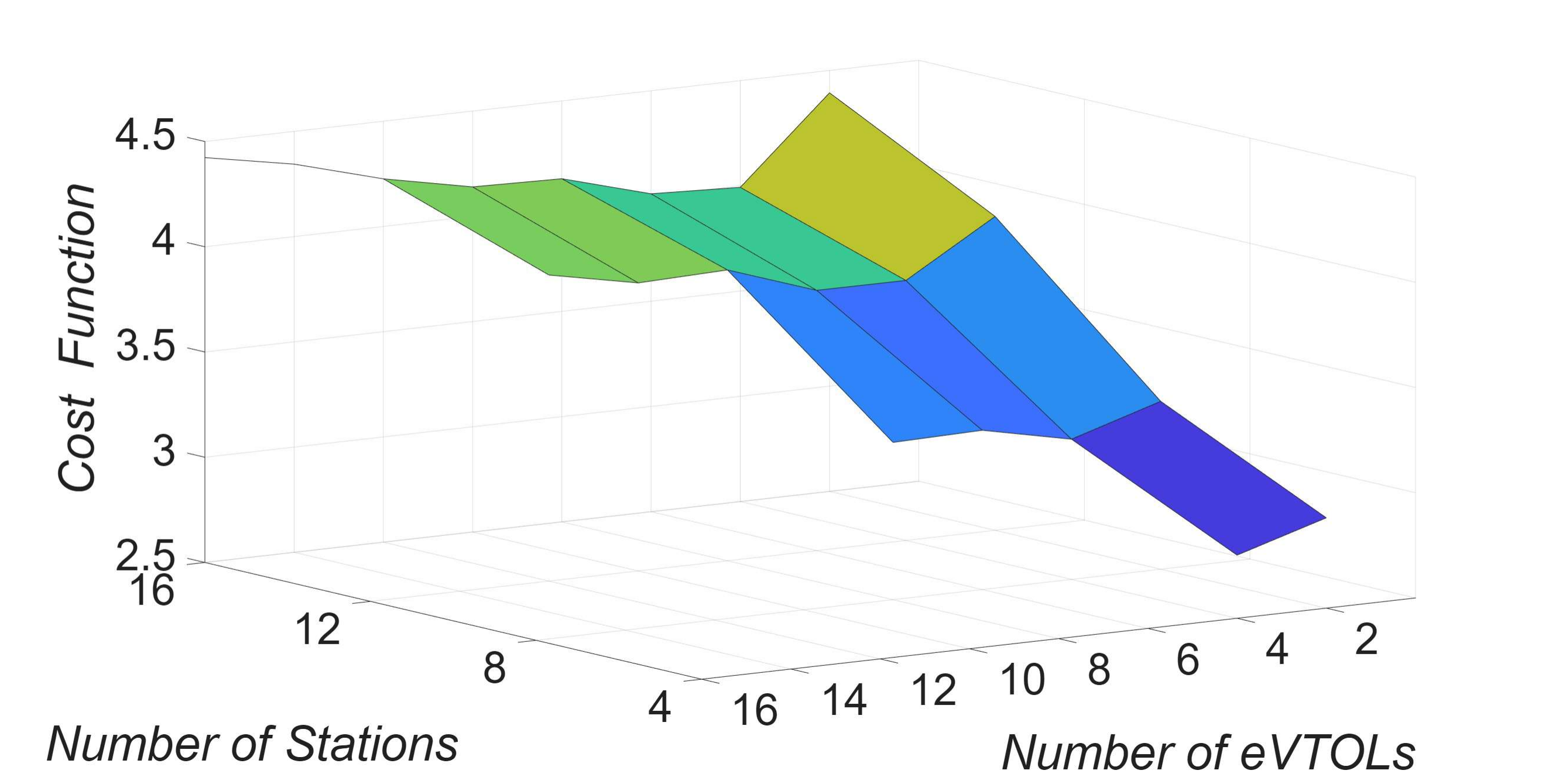}
        \caption{$D=100$}
        \label{fig:cost_100}
    \end{subfigure}
    \hfill
    \begin{subfigure}[t]{0.45\textwidth}
        \centering
        \includegraphics[width=0.94\linewidth]{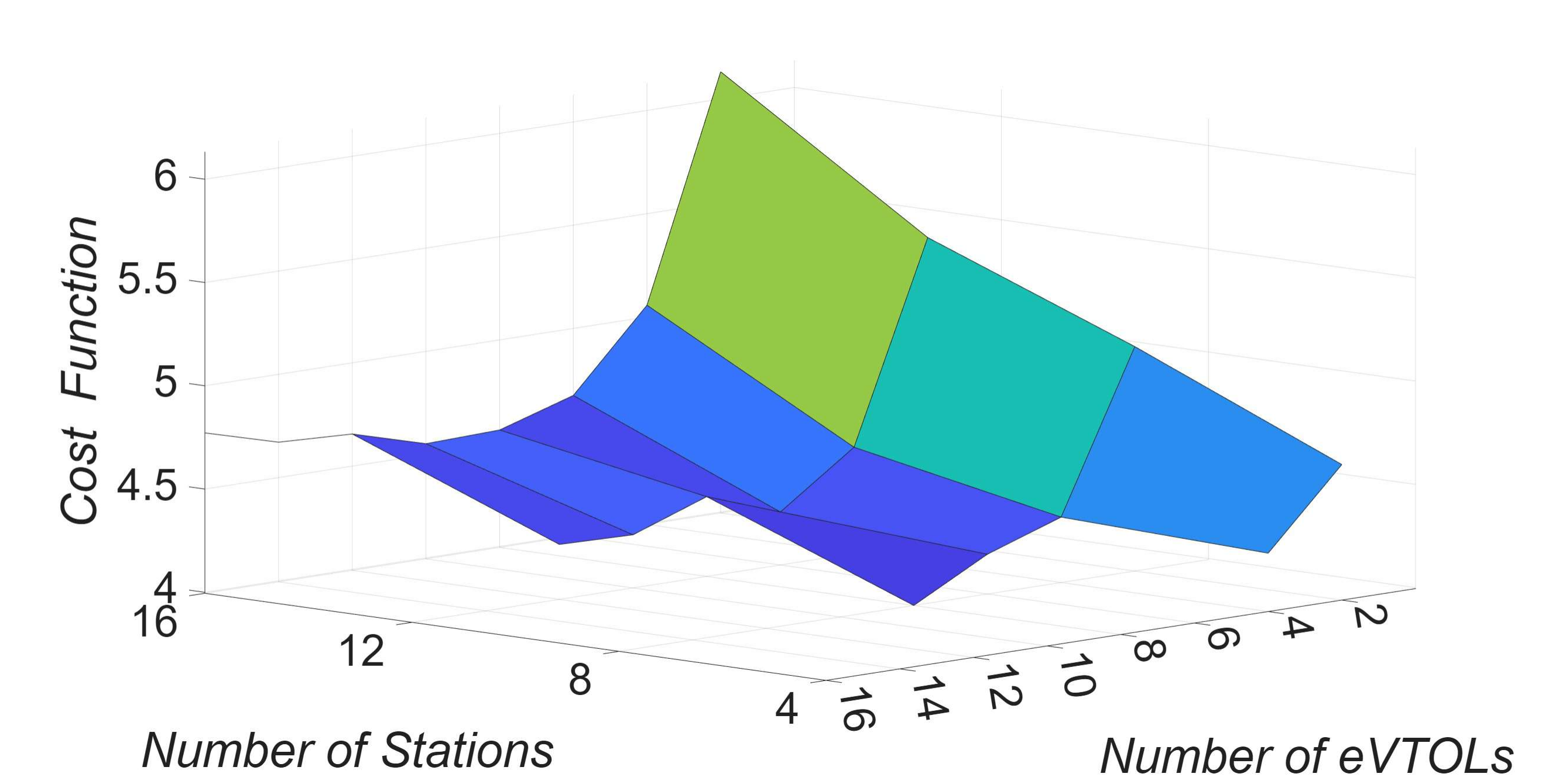}
        \caption{$D=500$}
        \label{fig:cost_500}
    \end{subfigure}

    \vspace{2mm}

    \begin{subfigure}[t]{0.45\textwidth}
        \centering
        \includegraphics[width=0.94\linewidth]{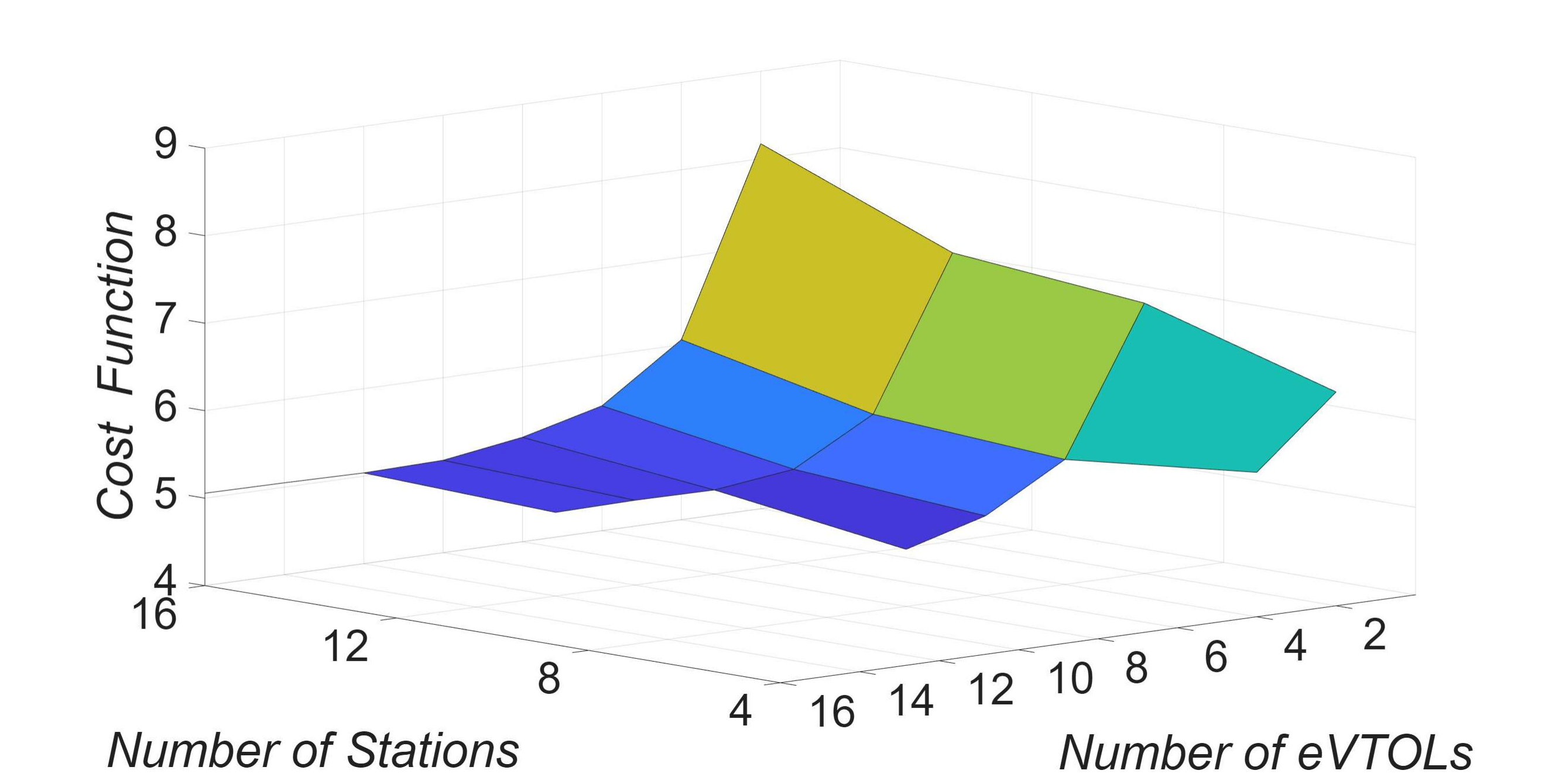}
        \caption{$D=1000$}
        \label{fig:cost_1000}
    \end{subfigure}
    \hfill
    \begin{subfigure}[t]{0.45\textwidth}
        \centering
        \includegraphics[width=0.94\linewidth]{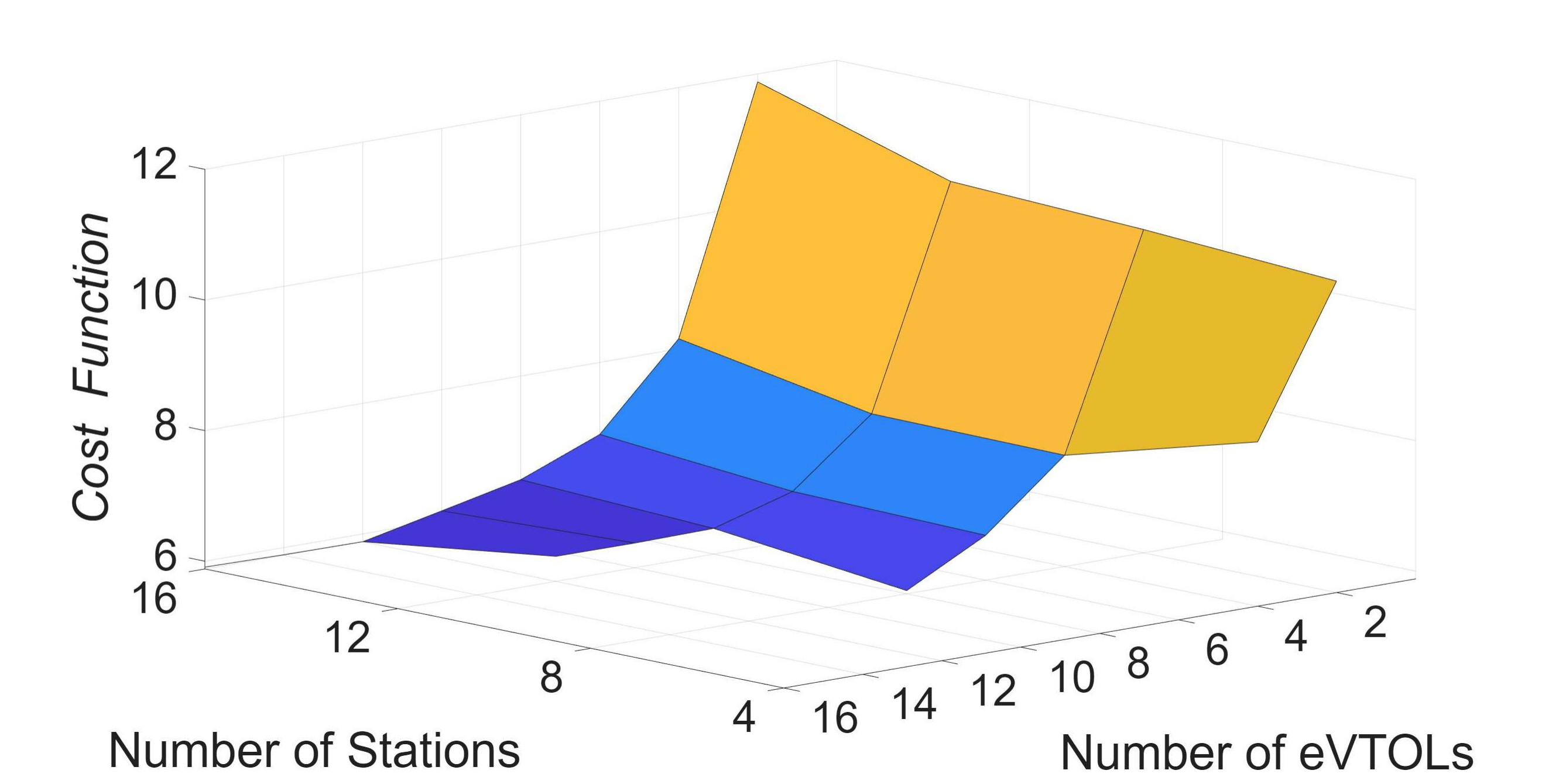}
        \caption{$D=2000$}
        \label{fig:cost_2000}
    \end{subfigure}

    \caption{Objective landscape for the weighted cost function $J$ as a function of station count $K$ and fleet size $F$ under four demand scenarios: (a) $D=100$, (b) $D=500$, (c) $D=1000$, and (d) $D=2000$}
    \label{fig:cost_surfaces}
\end{figure*}

Fig.~\ref{fig:cost_surfaces} shows the cost function $J$ evaluated over the tested $(K,F)$ grid for $D\in\{100,500,1000,2000\}$. The low-cost region shifts toward larger $(K,F)$ as demand increases, consistent with Table~\ref{tab:opt_networks}. The selected designs fall within these low-cost basins, indicating limited sensitivity to nearby $(K,F)$ choices within the candidate set.

\begin{figure*}[!t]
    \centering
    \begin{subfigure}[t]{0.45\textwidth}
        \centering
        \includegraphics[width=0.94\linewidth]{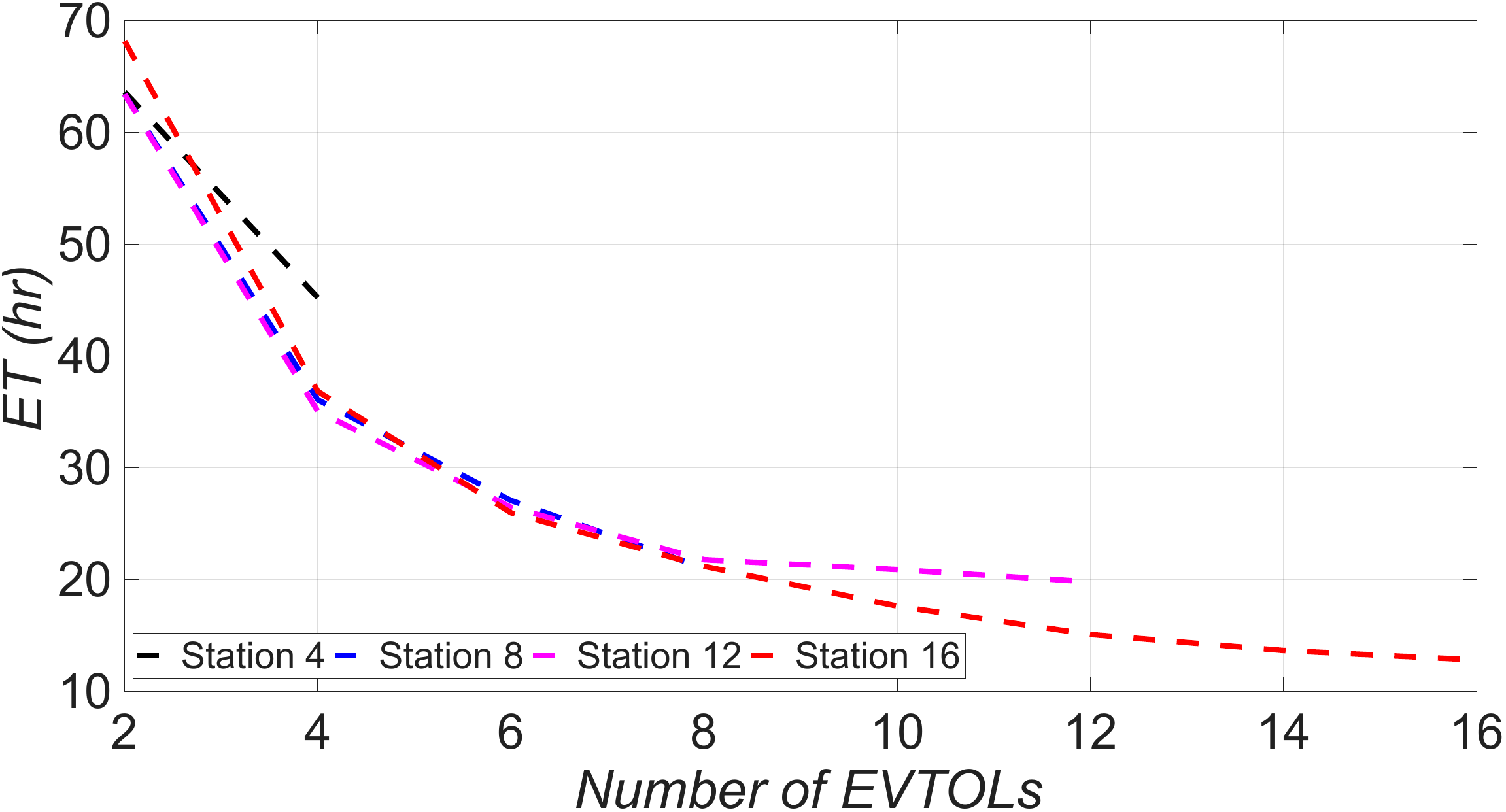}
        \caption{$\mathrm{ET}$ vs.\ $F$}
        \label{fig:tradeoff_2000_et}
    \end{subfigure}
    \hfill
    \begin{subfigure}[t]{0.45\textwidth}
        \centering
        \includegraphics[width=0.94\linewidth]{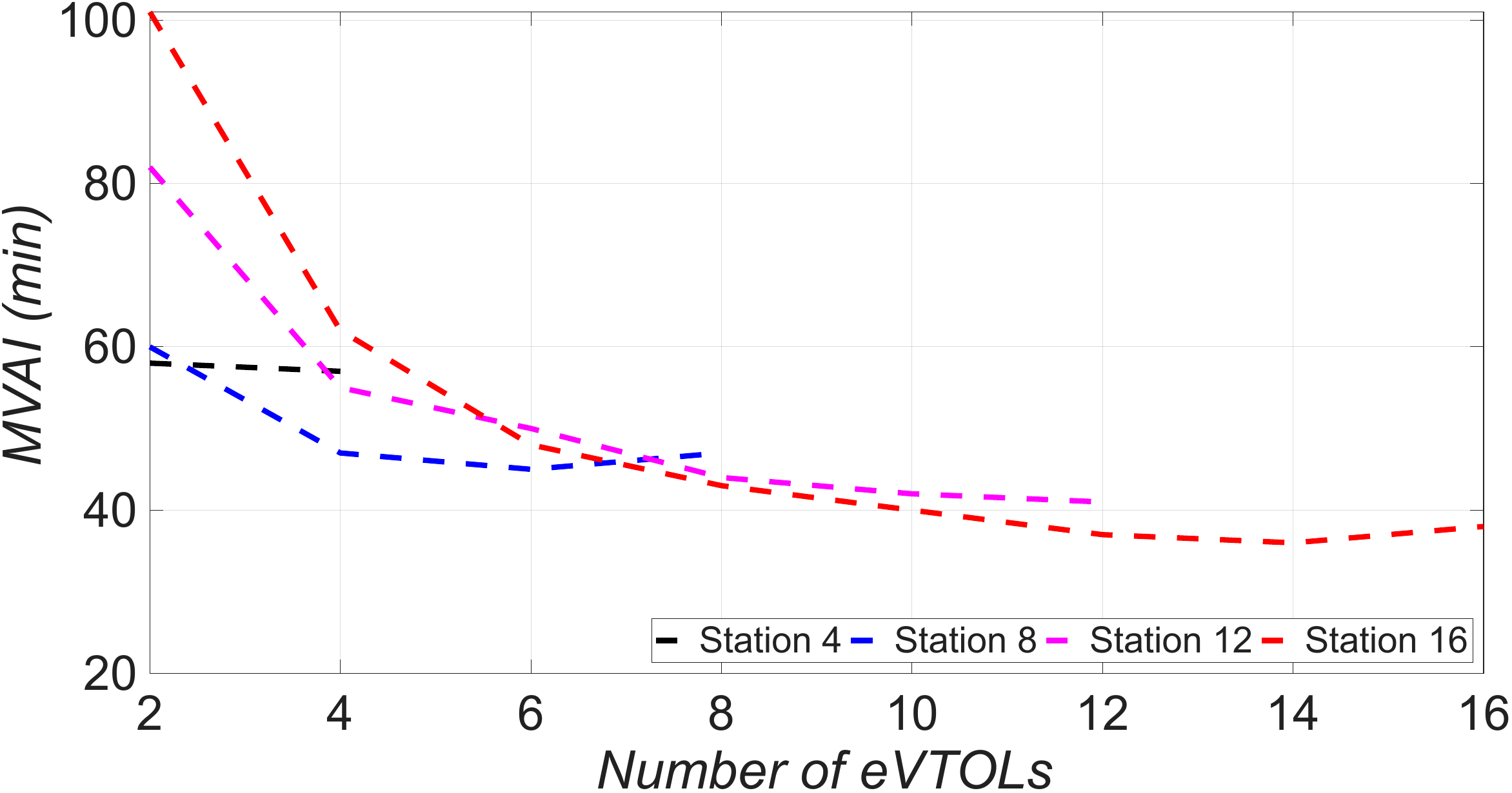}
        \caption{$\mathrm{MVAI}$ vs.\ $F$}
        \label{fig:tradeoff_2000_mv}
    \end{subfigure}

    \vspace{2mm}

    \begin{subfigure}[t]{0.45\textwidth}
        \centering
        \includegraphics[width=0.94\linewidth]{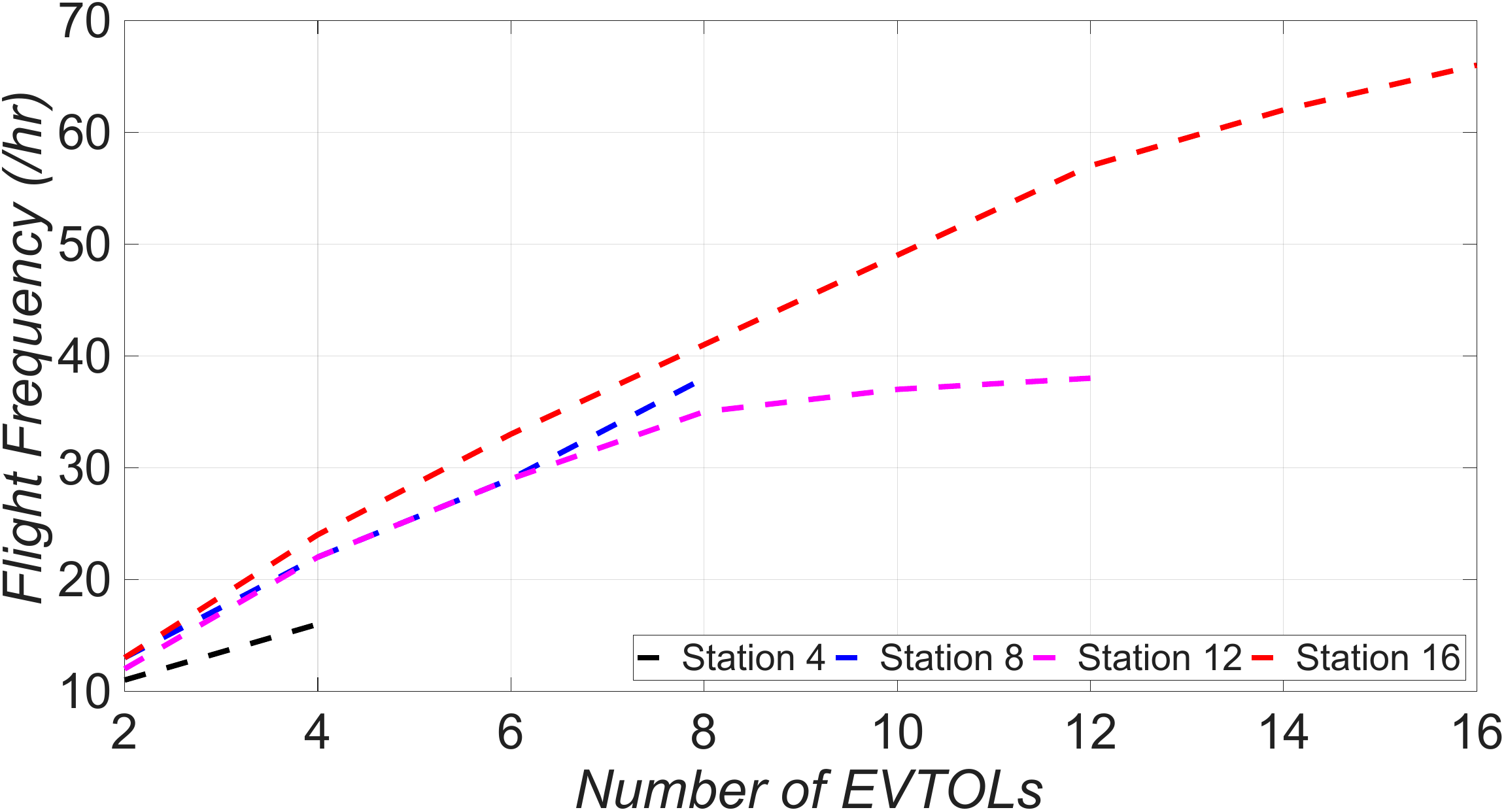}
        \caption{Flight frequency vs.\ $F$}
        \label{fig:tradeoff_2000_freq}
    \end{subfigure}
    \hfill
    \begin{subfigure}[t]{0.45\textwidth}
        \centering
        \includegraphics[width=0.94\linewidth]{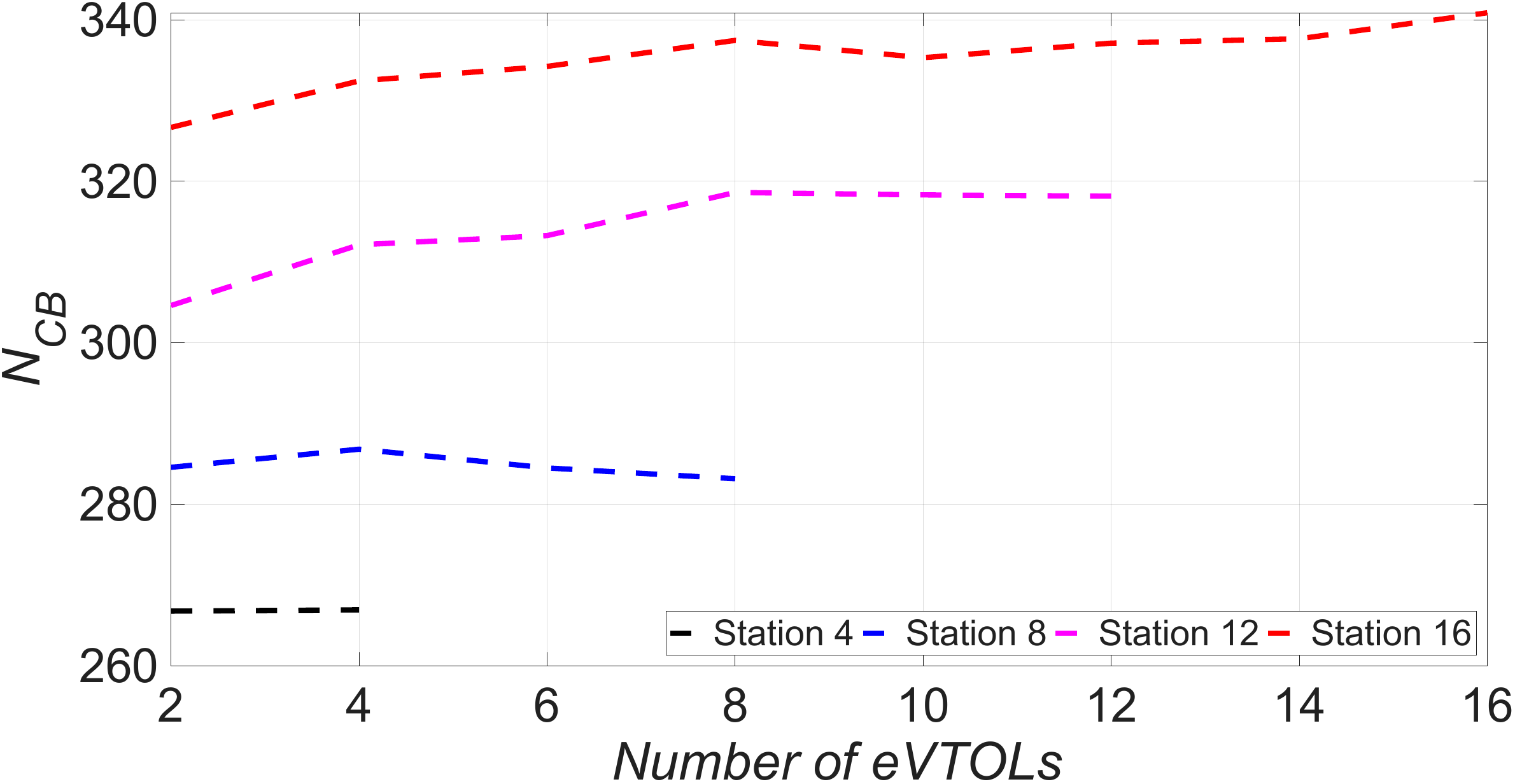}
        \caption{Battery swaps $\mathrm{CB}$ vs.\ $F$}
        \label{fig:tradeoff_2000_ncb}
    \end{subfigure}

    \caption{Station--fleet tradeoffs for $D=2000$ passengers. Panels report DES outcomes across tested station counts $K$ and fleet sizes $F$: (a) completion/evacuation time, (b) service regularity proxy, (c) flight frequency, and (d) battery swaps}
    \label{fig:tradeoff_2000_abcd}
\end{figure*}

\begin{figure}[tb]
    \centering
    \includegraphics[width=\columnwidth]{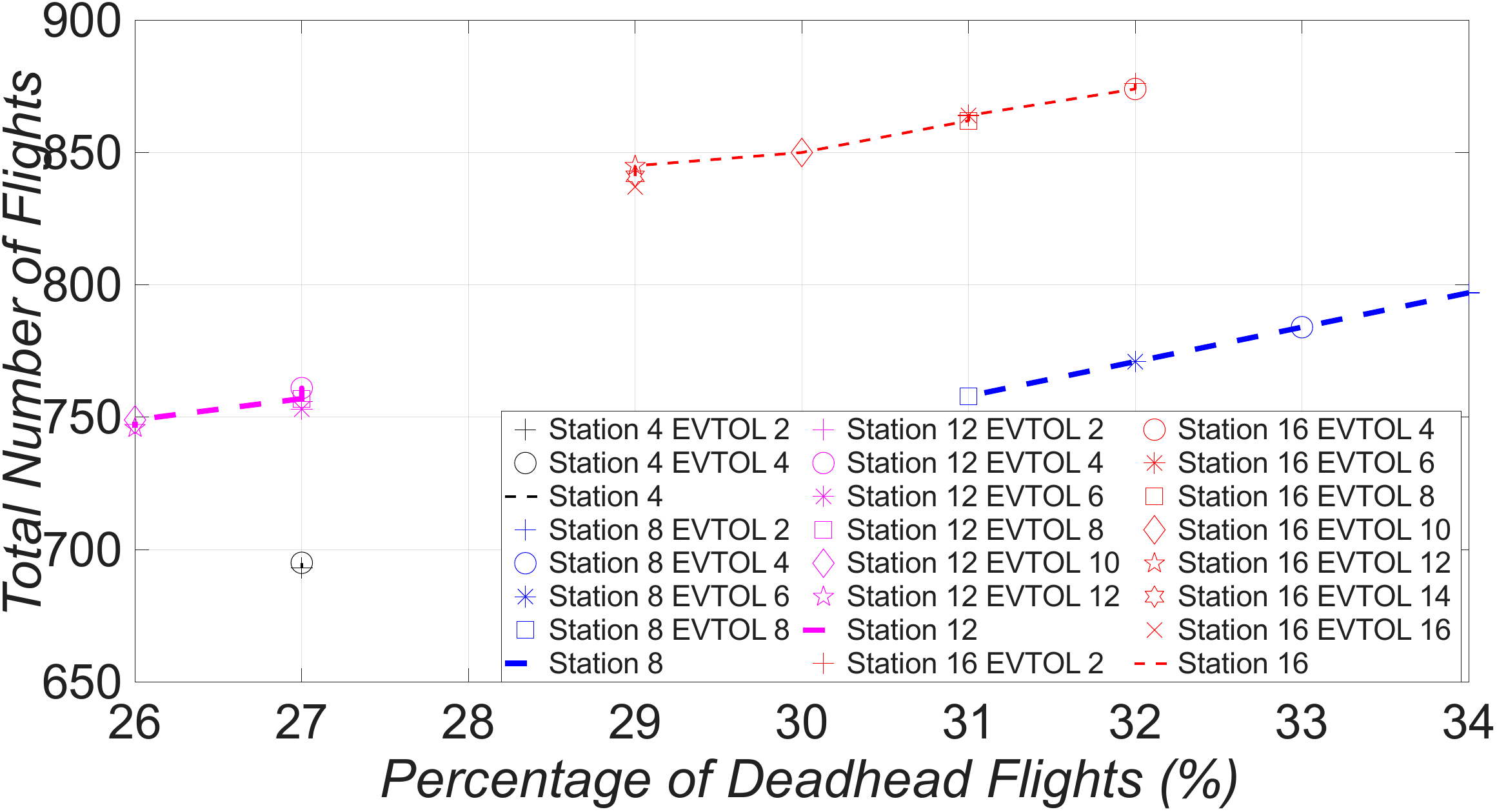}
    \caption{Deadhead burden for $D=2000$: total number of flights versus percentage of deadhead flights}
    \label{fig:dhr_totalflights_2000}
\end{figure}

For the $D=2000$ scenario, Fig.~\ref{fig:tradeoff_2000_abcd} breaks down how performance changes with fleet size $F$ for each station count $K$. As expected, for any fixed station configuration, increasing $F$ reduces the completion/evacuation time $\mathrm{ET}$ (Fig.~\ref{fig:tradeoff_2000_et}) and improves service regularity, shown by a decreasing mean vehicle arrival interval $\mathrm{MVAI}$ (Fig.~\ref{fig:tradeoff_2000_mv}). At the same time, the realized flight frequency increases with $F$ (Fig.~\ref{fig:tradeoff_2000_freq}), indicating more frequent service as more vehicles are available. This improvement comes with higher operational load; the number of battery swaps $\mathrm{CB}$ generally increases with $F$ (Fig.~\ref{fig:tradeoff_2000_ncb}), reflecting higher total energy use under more intensive operations. Fig.~\ref{fig:dhr_totalflights_2000} shows that deadheading does not disappear; while the total number of flights changes across $(K,F)$, the deadhead ratio remains nonzero, meaning empty repositioning flights are still needed to move vehicles toward remaining demand and keep the network balanced. This indicates that deadheading is not only a capacity issue but also a spatial-balancing issue. Even with different station--fleet combinations, vehicles must be repositioned because trip origins and destinations are not evenly distributed across the network.

\subsection{Door-to-Door Travel-Time Savings Feasibility}
\label{subsec:results_tts}

\begin{figure}[!t]
    \centering
    \includegraphics[width=\columnwidth]{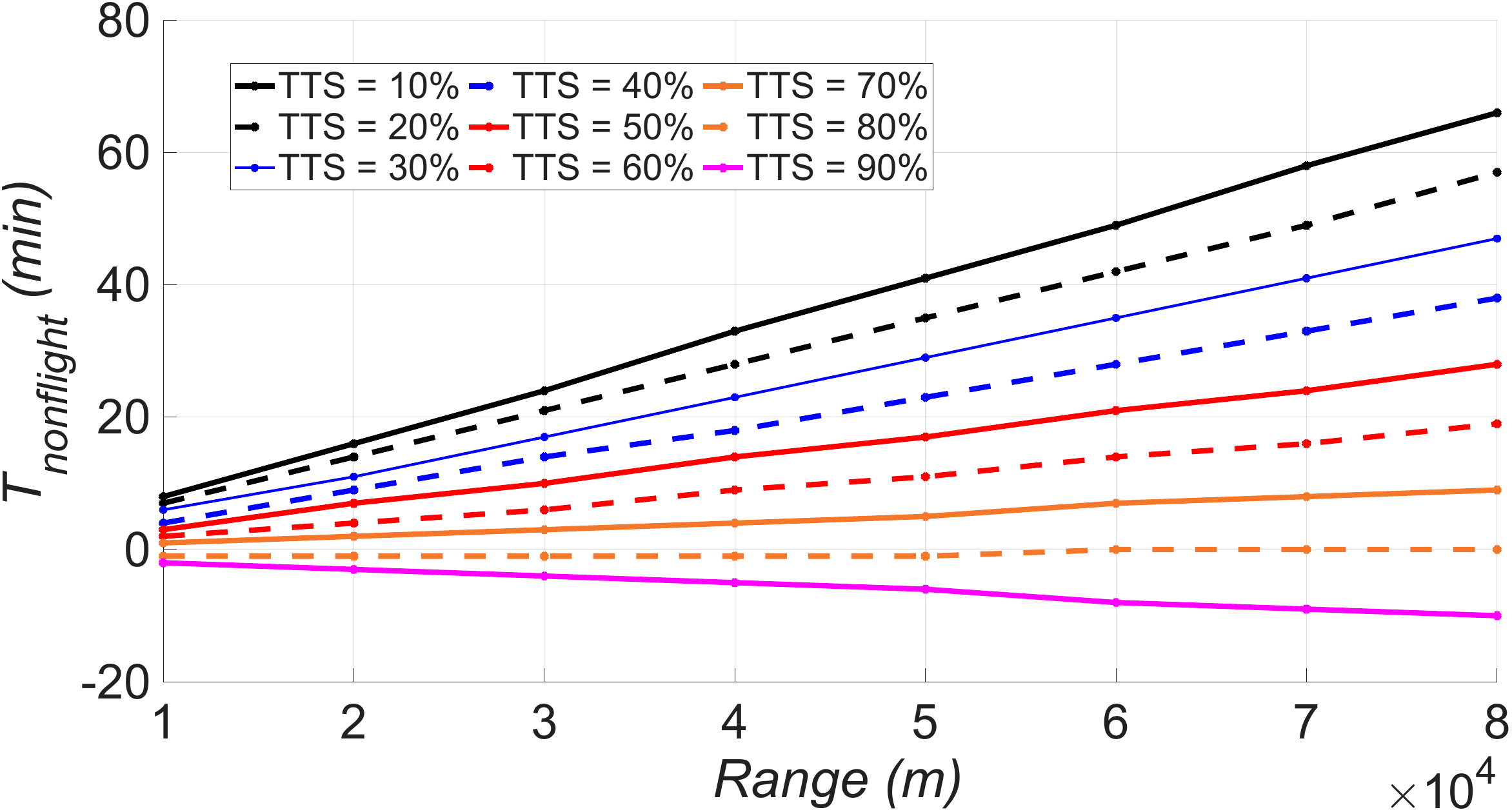}
    \vspace{-2mm}
    \caption{Maximum allowable non-flight time budget vs.\ station-to-station range for multiple travel-time savings targets at $v=80~\mathrm{km/h}$}
    \label{fig:tts_curve}
    \vspace{-2mm}
\end{figure}

Fig.~\ref{fig:tts_curve} shows the maximum allowable non-flight time budget at $v=80~\mathrm{km/h}$ for multiple travel-time savings targets. The budget decreases as the savings target increases and generally grows with range because ground travel time increases faster than flight time under the assumed routing factor. For a $40\%$ savings target, the allowable non-flight time is $26$ and $11$ min at $R=20$ km for $v=50$ and $80$ km/h, respectively, and increases to $71$ and $28$ min at $R=60$ km. Thus, UAM is more defensible for longer or congestion-heavy trips where sufficient non-flight time remains after accounting for flight time. As a sanity check, a representative Santa Clarita--LAX station pair has an inter-station distance of approximately $57$ km and an eVTOL flight time of about $14$ min. Google Maps reported a ground travel time of $65$--$120$ min for a nearby trip at 8:00 A.M. on March~2,~2026. For a $40\%$ savings target, this implies an allowable non-flight budget of approximately $[20,52]$ min after subtracting flight time and a $5$ min access time, consistent with Fig.~\ref{fig:tts_curve}.

\section{CONCLUSION AND FUTURE WORK}
\label{sec:conclusion}

This paper developed a demand-driven framework for on-demand Urban Air Mobility (UAM) network design that links demand construction, vertiport siting, discrete-event operational evaluation, and door-to-door travel-time savings analysis. Demand was estimated from public commuter and passenger activity data, candidate vertiport locations were generated using $K$-means clustering under range and spacing constraints, and network performance was evaluated using a DES with on-demand dispatch, deadhead relocation, and an eVTOL performance model.

The Greater Los Angeles case study showed that the preferred station--fleet configuration increases with demand, from a four-station, four-vehicle network at low demand to a sixteen-station, twelve-vehicle network at the highest tested demand level. Increasing fleet size reduced completion time and improved vehicle-arrival regularity, but did not eliminate deadhead flights, indicating that empty repositioning is driven by both fleet availability and spatial OD imbalance.

The results should be interpreted within several limitations: the Monte Carlo OD generation is an operational stress test rather than a calibrated empirical OD model, $K$-means is used as a scalable siting baseline rather than a globally optimal facility-location method, and the DES does not explicitly model passenger arrivals, queues, vertiport capacity, alternative relocation policies, weight sensitivity, or airspace conflicts.

Future work will address these limitations by perturbing the empirical OD matrix induced by the spatial demand model, extending siting through hub-location and facility-location optimization with demographic, land-use, equity, and regulatory constraints, and expanding the operations model to include queueing, vertiport capacity, vehicle-routing-based scheduling, alternative dispatch policies, weight-sensitivity analysis, and airspace conflict constraints.





\balance
\bibliographystyle{IEEEtran}
\bibliography{references}

\end{document}